\documentclass[letterpaper, 10pt, journal, twoside]{IEEEtran}

\IEEEoverridecommandlockouts                              

\usepackage{multicol}
\usepackage[hidelinks]{hyperref}
\usepackage{bm}
\usepackage{url}
\usepackage{graphicx} 
\usepackage{mathptmx} 
\usepackage[mathscr]{euscript}  
\usepackage{amsmath} 
\usepackage{amssymb}  
\usepackage{balance}
\usepackage{multirow}
\usepackage[load=prefixed]{siunitx}
\usepackage{xcolor}
\usepackage{adjustbox}
\usepackage{array}
\usepackage{multirow}
\usepackage{booktabs}
\let\labelindent\relax
\usepackage[inline]{enumitem}
\usepackage[hang,flushmargin]{footmisc}
\usepackage{microtype}
\usepackage[resetlabels]{multibib}
\newcites{New}{References}
\usepackage{cuted}
    
\definecolor{sky}{RGB}{70, 130, 180}
\definecolor{road}{RGB}{128, 64, 128}
\definecolor{sidewalk}{RGB}{244, 35, 232}
\definecolor{grass}{RGB}{152, 251, 152}
\definecolor{vegetation}{RGB}{107, 142, 35}
\definecolor{building}{RGB}{70, 70, 70}
\definecolor{pole}{RGB}{153, 153, 153}
\definecolor{dynamic}{RGB}{111, 74, 0}
\definecolor{other}{RGB}{220, 220, 0}

\newcommand{\secref}[1]{Sec.~\ref{#1}}

\renewcommand{\eqref}[1]{Eq.~(\ref{#1})}
\newcommand{\figref}[1]{Fig.~\ref{#1}}
\newcommand{\tabref}[1]{Tab.~\ref{#1}}

\DeclareMathAlphabet\mathbfcal{OMS}{cmsy}{b}{n}
\DeclareSIUnit\degm{deg/m}
\DeclareMathOperator*{\OPLUS}{\oplus}

\newcolumntype{P}[1]{>{\centering\arraybackslash}p{#1}}
\newcommand\crule[3][black]{\textcolor{#1}{\rule{#2}{#3}}}

\renewcommand{\baselinestretch}{0.95}

\newcolumntype{R}[2]{%
    >{\adjustbox{angle=#1,lap=\width-(#2)}\bgroup}%
    l%
    <{\egroup}%
}

\title{\LARGE \bf
VLocNet++: Deep Multitask Learning for \\Semantic Visual Localization and Odometry
}

\author{Noha Radwan$^*$ \and Abhinav Valada$^*$ \and Wolfram Burgard
\thanks{$^*$These authors contributed equally. All authors are with the Department of Computer Science, University of Freiburg, Germany.}
}

\begin{document}

\onecolumn
{\Large

\noindent\textcopyright IEEE. Personal use of this material is permitted. Permission from IEEE must be obtained
for all other uses, in any current or future media, including reprinting/republishing this material
for advertising or promotional purposes, creating new collective works, for resale or redistribution
to servers or lists, or reuse of any copyrighted component of this work in other works.\\

%
\noindent{Pre-print of the article that will appear in the\\ \textbf{IEEE Robotics and Automation Letters (RA-L)}.}\\

\noindent{Please cite this paper as:}\\
N. Radwan, A. Valada, W. Burgard, "VLocNet++: Deep Multitask Learning for Semantic Visual Localization and Odometry", \textit{IEEE Robotics and Automation Letters (RA-L)}, 3(4):4407-4414, 2018.\\

\noindent{BibTex:}\\
\\
@article$\lbrace$radwan18ral,\\
author = $\lbrace$Noha Radwan and Abhinav Valada and Wolfram Burgard$\rbrace$,\\
title = $\lbrace$VLocNet++: Deep Multitask Learning for Semantic Visual Localization and Odometry$\rbrace$,\\
journal = $\lbrace$IEEE Robotics and Automation Letters (RA-L)$\rbrace$,\\
volume = $\lbrace$3$\rbrace$,\\
number = $\lbrace$4$\rbrace$,\\
pages = $\lbrace$4407-4414$\rbrace$,\\
year = $\lbrace$2018$\rbrace$,\\
doi = $\lbrace$10.1109/LRA.2018.2869640$\rbrace$\\
$\rbrace$
}
\twocolumn

\maketitle
\thispagestyle{empty}
\pagestyle{empty}

\begin{abstract}
Semantic understanding and localization are fundamental enablers of robot autonomy that have for the most part been tackled as disjoint problems. While deep learning has enabled recent breakthroughs across a wide spectrum of scene understanding tasks, its applicability to state estimation tasks has been limited due to the direct formulation that renders it incapable of encoding scene-specific constrains. In this work, we propose the VLocNet++ architecture that employs a multitask learning approach to exploit the inter-task relationship between learning semantics, regressing 6-DoF global pose and odometry, for the mutual benefit of each of these tasks. Our network overcomes the aforementioned limitation by simultaneously embedding geometric and semantic knowledge of the world into the pose regression network. We propose a novel adaptive weighted fusion layer to aggregate motion-specific temporal information and to fuse semantic features into the localization stream based on region activations. Furthermore, we propose a self-supervised warping technique that uses the relative motion to warp intermediate network representations in the segmentation stream for learning consistent semantics. Finally, we introduce a first-of-a-kind urban outdoor localization dataset with pixel-level semantic labels and multiple loops for training deep networks. Extensive experiments on the challenging Microsoft 7-Scenes benchmark and our DeepLoc dataset demonstrate that our approach exceeds the state-of-the-art outperforming local feature-based methods while simultaneously performing multiple tasks and exhibiting substantial robustness in challenging scenarios.
\end{abstract}
\vspace{-0.3cm}
\section{Introduction}
\label{sec:intro}

Autonomous robots today are a complex ensemble of modules each of which specializes in a particular domain such as state estimation and scene understanding. While significant strides have been made considering these domains separately~\cite{sattler2017benchmarking, valada2017adapnet, melekhov2017cnnBspp}, very little progress has been made towards exploiting the relationship between them. In this work, we focus on jointly learning three diverse vital tasks that are crucial for robot autonomy, namely, semantic segmentation, visual localization and odometry from consecutive monocular images. We approach this problem from a multitask learning (MTL) perspective with the goal of learning more accurate localization and semantic segmentation models by leveraging the predicted ego-motion. This problem is extremely challenging as it involves simultaneously learning cross-domain tasks that perform pixel-wise classification and regression with different units and scales. However, this joint formulation enables inter-task learning which improves both generalization capabilities and alleviates the problem of requiring vast amounts of labeled training data which is especially hard to obtain in the robotics domain. Moreover, as robots are equipped with limited resources, a joint model is more efficient for deployment and enables real-time inference on a consumer grade GPU.

Most existing CNN-based metric localization approaches~\cite{kendall2017geometric,wu2017,naseer17iros} perform direct pose regression from image embeddings using naive loss functions. In order to more effectively encode knowledge about the environment, we propose a principled approach to embed geometric and semantic knowledge into the pose regression model. Our network utilizes our Geometric Consistency loss function~\cite{valada18icra} that incorporates relative motion information to learn a model that is globally consistent. Firstly, unlike the previous approach~\cite{valada18icra}, to efficiently utilize the learned motion specific features from the previous timestep, we employ an adaptive weighting technique to aggregate motion-specific temporal information. Secondly, by jointly estimating the semantics, we instill structural cues about the environment into the pose regression network and implicitly pull the attention towards more informative regions in the scene. Existing semantics-aware localization techniques extract predefined stable features, emphasize~\cite{kobyshev2014} or combine them with local features~\cite{Singh2016} but often fail when the predefined structures are occluded or not visible in the scene. Our approach is robust to such situations as it uses our proposed adaptive fusion layer to fuse learned relevant features not only based on the semantic category but also the activations in the region. 

\begin{figure}
\centering
\includegraphics[width=0.99\linewidth]{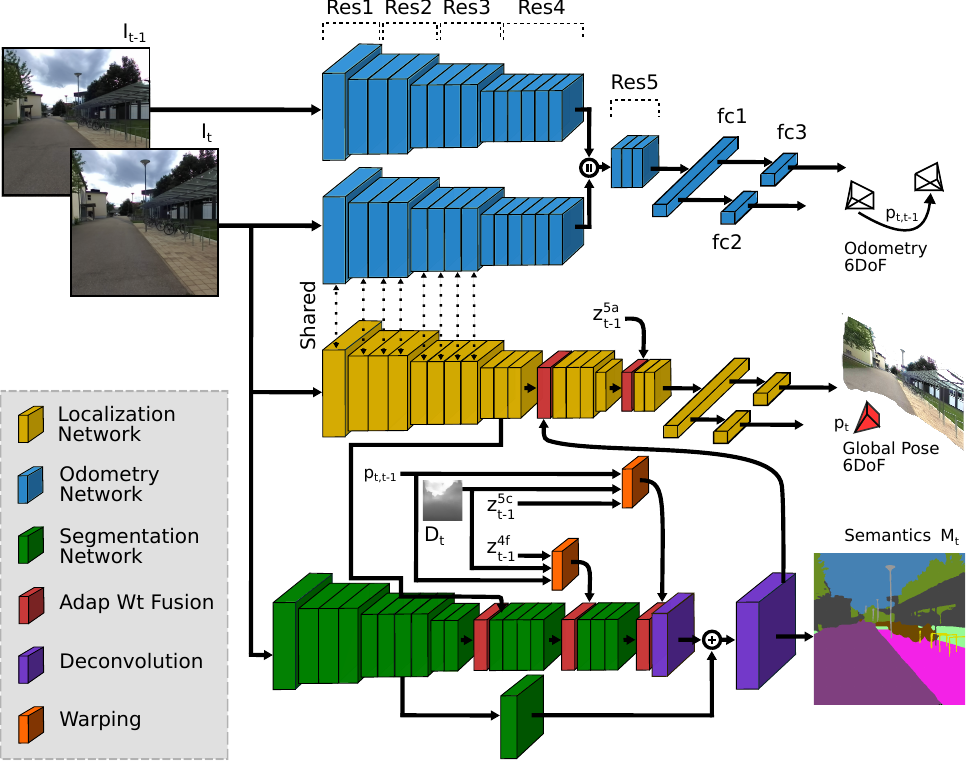}
\vspace{-0.6cm}
\caption{Schematic representation of our proposed VLocNet++ architecture. The network takes two consecutive monocular images ($I_t, I_{t-1}$) as input and simultaneously predicts the global 6-DoF pose $p_t$, odometry $p_{t,t-1}$ and semantics $M_t$ of the scene. $z_{t-1}^l$ denotes the feature maps of layer $l$ from the previous timestep and $D_t$ denotes a predicted depth map that is used for representational warping in the semantic stream.}
\label{fig:vlocnet++}
\vspace{-0.6cm}
\end{figure}


Predicting consistent semantics is a critical prerequisite for semantic visual localization. Inspired by early cognitive studies in humans showing the importance of learning self-motion for acquiring basic perceptual skills~\cite{nancy1980}, we propose a novel self-supervised semantic context aggregation technique leveraging the predicted relative motion from the odometry stream of our network. Using pixel-wise depth predictions from a CNN~\cite{MIFDB16} and differential warping, we fuse intermediate network representations from the previous timestep into the current frame using our proposed adaptive weighted fusion layer. This enables our semantic segmentation network to aggregate more scene-level context, thereby improving the performance and leading to faster convergence.

In summary, the primary contributions of this paper are as follows:
\begin{enumerate*}[label=(\roman*)]
  \item A novel MTL framework for jointly learning semantics, visual localization and odometry from consecutive monocular images.
  \item A CNN architecture for pose regression that significantly outperforms state-of-the-art approaches on the challenging Microsoft 7-Scenes benchmark.
  \item A self-supervised context aggregation technique based on differential warping that improves semantic segmentation and reduces the training time by half.
  \item A novel adaptive weighted fusion layer for element-wise fusion of feature maps based on region activations to exploit inter/intra task dependencies. 
  \item Finally, to facilitate this work, we introduce a first-of-a-kind outdoor dataset consisting of multiple loops with pixel-level semantic labels and localization ground truth. It contains repetitive, translucent and reflective surfaces, weakly 
textured regions and low-lighted scenes with shadows, thereby making it 
extremely challenging for benchmarking a variety of tasks.
\end{enumerate*}
\vspace{-0.05cm}
\section{Related Works}
\label{sec:relatedWorks}

Over the past decade there has been a gradual shift from employing traditional handcrafted pipelines to learning-based methods particularly for perception
related tasks.
In this section, we discuss recent learning-based approaches for 
multitask learning, pose regression and semantic segmentation.

\noindent\textbf{Multitask Learning} can be defined as an inductive transfer 
mechanism that improves generalization by leveraging domain specific information 
from related tasks~\cite{Caruana1997}. It has been applied to a wide range of 
tasks~\cite{teichmann2016multinet}. 
Bilen~\textit{et al.} propose the use of 
an instance normalization layer to train a network that recognizes objects across multiple visual domains including 
digits, signs and faces~\cite{bilenV17}.
In~\cite{shazeerMMDLHD17}, the 
authors introduce a model with a sparsely-gated mixture of experts layer 
 for the task of language modeling and 
machine translation. Multinet~\cite{teichmann2016multinet} proposes a unified 
architecture consisting of a shared encoder and task-specific decoders for 
classification, detection and segmentation. For combining different loss 
functions in a multitask model, Kendall~\textit{et al.}~\cite{kendall2017multi} 
propose a loss function based on maximizing the Gaussian likelihood using 
homoscedastic task uncertainty. While the aforementioned approaches mostly have shared
parts of the network that learn low-level features followed by individual task-specific branches,
 we propose a novel adaptive 
weighted fusion layer that learns the most favorable weighting of feature maps for the 
mutual benefit of the tasks.

\noindent\textbf{Visual Localization} has been addressed by a variety of approaches ranging from image retrieval~\cite{kalantidis2011viral}, local feature-based pipelines~\cite{sattler2017benchmarking}, to end-to-end learning methods~\cite{BrachmannKNSMGR16}.
Recently, pre-trained DCNNs designed for 
classification have been successfully adapted for pose regression. 
Kendall~\textit{et al.} proposed PoseNet~\cite{kendall2015convolutional}, an end-to-end approach for directly regressing 
the 6-DoF camera pose from a monocular image using a DCNN. Since then several improvements have 
been proposed in terms of 
incorporating Long-Short Term Memory (LSTM) 
units for dimensionality reduction~\cite{walch16spatialstms}, symmetric 
encoder-decoder architecture for regression~\cite{MelekhovYKR17} and an improved 
loss function based on scene geometry~\cite{kendall2017geometric}. 
NNnet~\cite{laskar2017camera} employs a hybrid approach where a DCNN trained on 
relative camera pose estimation is employed to extract features for identifying the nearest neighbors of a query image among the database images. 
Recently, Brachmann~\textit{et al.} proposed a differentiable version of 
RANSAC (DSAC)~\cite{BrachmannKNSMGR16} and a successor 
version~\cite{brachmann2017} for camera localization. Currently this approach~\cite{brachmann2017} achieves state-of-the-art performance 
on the Microsoft 7-Scenes benchmark. More recently, 
VLocNet~\cite{valada18icra} presented the Geometric Consistency Loss function 
that constricts the search space with the relative motion information during 
training to obtain pose estimates that are consistent with the true motion 
model. The focus of this paper is to improve VLocNet's performance by adaptively 
fusing semantic features and aggregated motion-specific information from the 
previous timestep into the localization network.

\noindent\textbf{Visual Odometry:} Another similar line of work is to estimate 
the incremental change in position from images. Nicolai~\textit{et 	al.}~\cite{NicolaiRSSw2016} employed a simple Siamese architecture with 
alternating convolution and pooling layers to estimate the transforms from 
consecutive point clouds. Konda~\textit{et al.}~\cite{Konda2015LearningVO} 
proposed an end-to-end architecture for learning ego-motion from a sequence of 
RGB-D images using a prior set of discretized velocities and directions. 
DeepVO~\cite{MohantyDeepVO} presents an AlexNet-based Siamese architecture for 
odometry estimation from monocular images, in which they also experiment with 
appending FAST features along with the images as input to the network. 
Melekhov~\textit{et al.}~\cite{melekhov2017cnnBspp} propose a CNN architecture 
that incorporates spatial pyramid pooling and demonstrates improved performance 
compared to local feature-based approaches that utilize SIFT or ORB.

\noindent\textbf{Semantic Segmentation:} Fully Convolutional Neural Networks 
(FCNs)~\cite{long2015} first proposed the encoder-decoder model replacing inner-product layers with convolution layers to 
enable pixel-wise classification. Several networks built upon FCNs by 
introducing more refinement stages
~\cite{oliveira2016}, efficient non-linear upsampling 
schemes~\cite{kendall2015}, adding global context~\cite{liu2015} and pyramid 
pooling for context aggregation~\cite{zhao2017pspnet}. Yu~\textit{et 
al.}~\cite{YuKoltun2016} proposed a context module the uses dilated convolutions 
to enlarge the receptive field. DeepLab~\cite{ChenPK0Y16} proposed using 
multiple parallel dilated convolutions with different sampling rates for 
multi-scale learning in addition to using CRFs for post-processing. 
\mbox{AdapNet~\cite{valada2017adapnet}} introduced multi-scale residual blocks 
with dilated convolutions as parallel convolutions to enable faster inference 
without compromising the performance. Ma~\textit{et al.}~\cite{lingni17iros} 
proposed warping nearby frames which do not have ground truth into the keypoint 
annotated frame for \mbox{RGB-D} semantic mapping. For learning consistent 
semantics in VLocNet++, we build upon AdapNet's model and propose a 
self-supervised warping technique for scene-level context aggregation. 
Our model for learning semantics in this work differs from Ma~\textit{et 
al.}~\cite{lingni17iros} as they warp feature maps from nearby frames to 
calculate the supervised loss when ground truth for those frames is absent. 
However, in this work we warp feature maps of the preceding frame and fuse them using the proposed adaptive fusion layer which leads to improved accuracy and faster convergence.
%
\vspace{-0.05cm}
\section{Technical Approach}
\label{sec:approach}

In this section, we detail our MTL framework for jointly estimating global pose, odometry and semantic segmentation from consecutive monocular images. While the approach presented in this paper focuses on joint learning of the aforementioned tasks, each of the task-specific models can be deployed independently during test-time. We propose a novel strategy for encoding geometric and structural constraints into the pose regression network, namely by incorporating information from the previous timesteps to accumulate motion specific information and by adaptively fusing semantic features based on the activations in the region using our proposed fusion scheme. As being able to predict robust semantics is an essential prerequisite for the proposed fusion, we present a new self-supervised warping technique for aggregating scene-level context in the semantic segmentation model. Our architecture, depicted in~\figref{fig:vlocnet++} consists of four CNN streams; a global pose regression stream, a semantic segmentation stream and a Siamese-type double stream for visual odometry estimation. 

Given a pair of consecutive monocular images $I_{t-1}, I_{t} \in \mathbb{R}^{\rho}$, the pose regression stream predicts the global pose $\mathbf{p}_{t}=[\mathbf{x}_{t}, \mathbf{q}_{t}]$ for image $I_{t}$, where $\mathbf{x} \in \mathbb{R}^3$ denotes the translation and $\mathbf{q} \in \mathbb{R}^4$ denotes the rotation in quaternion representation, while the semantic stream predicts a pixel-wise segmentation mask $M_t$ mapping each pixel $u$ to one of the $\mathit{C}$ semantic classes, and the odometry stream predicts the relative motion $\mathbf{p}_{t, t-1}=[\mathbf{x}_{t, t-1}, \mathbf{q}_{t, t-1}]$ between consecutive input frames. $z^l$ denotes the feature maps from layer $l$ of a particular stream. In the remainder of this section, we describe the constituting network components and our MTL scheme.
\vspace{-0.2cm}
\subsection{Geometrically Consistent Pose Regression}

Our model for regressing the global pose is based on the recently proposed VLocNet~\cite{valada18icra}  architecture. It has five residual blocks that downsample the feature maps by half at each block, similar to the full preactivation ResNet-50 architecture, but replaces the conventional Rectified Linear Units (ReLUs) activation function with Exponential Linear Units (ELUs), which help learning representations that are more robust to noise and also lead to faster convergence. We add a global average pooling layer after the fifth residual block, followed by three inner-product layers \textit{fc1, fc2 and fc3} of dimensions 1024, 3 and 4 respectively, where \textit{fc2} and \textit{fc3} regress the translational $\mathbf{x}$ and rotational $\mathbf{q}$ components of the pose. Unlike VLocNet which fuses the previous predicted pose directly using inner-product layers, we adopt a more methodological approach to provide the network with this prior. Fusing the previous prediction directly inhibits the network from being able to correlate motion specific spatial relations crucial for this task, with that of the previous timestep as the network does not retain these features thereafter. In this work, we integrate the network's intermediate representation $z_{t-1}^{5a}$ from the last downsampling stage (\textit{Res5a}) of the previous timestep using our proposed adaptive weighted fusion layer detailed in~\secref{sec:awf}. Our fusion scheme learns the most favorable element-wise weighting for this fusion, and when trained end-to-end with the Geometric Consistency Loss, enables aggregation of motion-specific features across the temporal dimension. We denote the aforementioned architecture as VLocNet$++_{\text{STL}}$ in our experiments.

As opposed to naively minimizing the Euclidean loss between the predicted poses and the ground truth, we employ the Geometric Consistency Loss function, which in addition to minimizing the Euclidean loss,  adds another loss term to constrain the current pose prediction by minimizing the relative motion error between the ground truth and the estimated motion from the odometry stream. By utilizing the predictions of the network from the previous timestep along with the current timestep, the relative motion loss term $\mathcal{L}_{Rel}\left(f\left(\theta \mid I_{t}\right)\right)$ can be computed as a weighted summation of the translational and rotational errors, where $\theta$ is defined to be the parameters of the network, and $f(\theta \mid I_t)$ denotes the predicted output of the network for image $I_t$. \eqref{eq:relLoss} details the relative motion loss term, in which we assume that the quaternion output of the network has been normalized a priori for ease of notation, and $\hat{s}_{x_{Rel}}, \hat{s}_{q_{Rel}}$ denote the learnable weighting variables for the translational and rotational components~\cite{kendall2017geometric}.
\begin{align}
\label{eq:relLoss}
\mathcal{L}_{Rel}\left(f\left(\theta \mid I_{t}\right)\right) &= \mathcal{L}_{x_{Rel}}\left(f\left(\theta \mid I_{t}\right)\right) \exp({-\hat{s}}_{x_{Rel}}) + \hat{s}_{x_{Rel}} \\ \nonumber &+ \mathcal{L}_{q_{Rel}}\left(f\left(\theta \mid I_{t}\right)\right) \exp({-\hat{s}}_{q_{Rel}}) + \hat{s}_{q_{Rel}} \\ \nonumber
\mathcal{L}_{x_{Rel}} \left(f\left(\theta \mid I_{t}\right)\right) &:= \left \| \mathbf{x}_{t, t-1} - \left(\hat{\mathbf{x}}_{t} - \hat{\mathbf{x}}_{t-1}\right) \right \|_2 \\ \nonumber
\mathcal{L}_{q_{Rel}} \left(f\left(\theta \mid I_{t}\right)\right) &:= \left \| \mathbf{q}_{t, t-1} - \left(\hat{\mathbf{q}}_{t-1}^{{-1}} \hat{\mathbf{q}}_{t} \right) \right \|_2.
\end{align}
Following the aforementioned notation, the Euclidean loss term can be defined as
\begin{align}
\label{eq:euclLoss}
\mathcal{L}_{Euc}\left(f\left(\theta \mid I_{t}\right)\right) &= \mathcal{L}_{x}\left(f\left(\theta \mid I_{t}\right)\right) \exp({-\hat{s}}_{x})\\ \nonumber &+ \hat{s}_{x} + \mathcal{L}_{q}\left(f\left(\theta \mid I_{t}\right)\right) \exp({-\hat{s}}_{q}) + \hat{s}_{q} \\ \nonumber
\mathcal{L}_{x}\left(f\left(\theta \mid I_{t}\right)\right) &:=  \left \| \mathbf{x}_{t} - \hat{\mathbf{x}}_{t}  
\right \|_2 \\ \nonumber
\mathcal{L}_{q}\left(f\left(\theta \mid I_{t}\right)\right) &:= \left
\| \mathbf{q}_{t} - {\hat{\mathbf{q}}_{t}}  \right \|_2.
\end{align}
The final loss term to be minimized is
\begin{align}
\label{eq:locLoss}
\mathcal{L}_{loc} \left(f\left(\theta \mid I_{t}\right)\right) &:=  \mathcal{L}_{Euc}\left(f\left(\theta \mid I_{t}\right)\right) + \mathcal{L}_{Rel} \left(f\left(\theta \mid I_{t}\right)\right).
\end{align}
By minimizing the aforementioned loss function, our network learns a model that is geometrically consistent with respect to the motion. Moreover, by employing a mechanism to aggregate motion specific features temporally, we enable the Geometric Consistency Loss to efficiently leverage this information.
\vspace{-0.2cm}
\subsection{Learning Visual Odometry}

Our proposed architecture for relative pose estimation takes a pair of consecutive monocular images $\left( I_{t-1}, I_{t} \right)$ as input and yields an estimate of ego-motion $\mathbf{p}_{t,t-1} = [ \mathbf{x}_{t,t-1}, \mathbf{q}_{t,t-1} ]$. We employ a dual-stream architecture in which each of the streams is identically similar in structure and is based on the full preactivation ResNet-50 model. We concatenate the feature maps of the individual streams before the last downsampling stage (end of \textit{Res4}) and convolve them through the last residual block, followed by an inner-product layer and two regressors for estimating the pose components. During training, we optimize the following loss function by minimizing the Euclidean error between the ground truth and the predicted motion.
\begin{align}\label{eq:voLoss}
 \mathcal{L}_{vo}\left(f\left(\theta \mid I_{t}, I_{t-1}\right)\right) := \mathcal{L}_{x}\left(f\left(\theta\mid I_{t}, I_{t-1}\right)\right) \exp({-\hat{s}}_{x_{vo}})\\ \nonumber + \hat{s}_{x_{vo}} + \mathcal{L}_{q}\left(f\left(\theta\mid I_{t}, I_{t-1}\right)\right) \exp({-\hat{s}}_{q_{vo}}) + \hat{s}_{q_{vo}},
\end{align}
where $\mathcal{L}_{x}$ and $\mathcal{L}_{q}$ refers to the translational and rotational components respectively. We also employ learnable weighting parameters to balance the scale between the translational and rotational components in the loss term. As shown in \figref{fig:vlocnet++}, the dual odometry streams have an architecture similar to the global pose regression network. In order to enable the inductive transfer of information between both networks, we share parameters between the odometry stream taking the current image $I_t$ and the global pose regression network as detailed in \secref{sec:awf}.\looseness=-1
\vspace{-0.1cm}
\subsection{Learning Semantics}
\label{sec:semantics}

Our model for learning consistent semantics has two variants: a single-task base architecture that takes a monocular image as input and predicts a pixel-wise segmentation mask (green and purple blocks in \figref{fig:vlocnet++}) and a multitask architecture built upon the base model that incorporates our proposed self-supervised warping and adaptive fusion layers (orange and red blocks).

\noindent\textbf{Network Architecture:} For the single-task base model, we adopt the AdapNet~\cite{valada2017adapnet} architecture which follows the general encoder-decoder design principle. Similar to the localization network, the encoder is based on the ResNet-50 model which includes skip connections and batch normalization layers that enable training such deep architectures by alleviating the vanishing gradient problem. The encoder learns highly discriminative semantic features and yields an output 16-times downsampled with respect to the input dimensions. While the decoder consists of two deconvolution layers and a skip convolution from the encoder for fusing high resolution feature maps and upsampling the downscaled feature maps back to the input resolution. The architecture also incorporates multi-scale ResNet blocks which have dilated convolutions parallel to the $3\times3$ convolutions for aggregating features from different spatial scales, concurrently maintaining fast inference times. 

Following the notation convention, we define a set of training images $\mathbfcal{T} = \{(I_n,M_n) \mid n = 1,\dots,N\}$, where $I_n=\{u_r \mid r=1,\dots,\rho\}$ denotes the input frame and the corresponding ground truth mask $M_n=\{m^n_r \mid r=1,\dots,\rho\}$, where $m^n_r \in \{1,\dots,C\}$ is the set of semantic classes. We define $\theta$ as the network parameters. Using the classification scores $s_j$ at each pixel $u_r$, we obtain a probabilities $\mathbf{P} = (p_1, \dots, p_C)$ with the $\mathsf{softmax}$ function $\sigma(.)$ such that
\begin{equation}
p_j(u_r,\theta \mid I_n) = \sigma\left(s_j\left(u_r, \theta\right)\right) = \frac{exp\left(s_j\left(u_r, \theta\right)\right)}{\sum^{C}_{k} exp\left(s_k\left(u_r,\theta\right)\right)}
\end{equation}
denotes the probability of pixel $u_r$ being classified with label $j$. The optimal $\theta$ is estimated by minimizing
\begin{equation}\label{eq:entropyloss}
\mathcal{L}_{seg}(\mathbfcal{T}, \theta) = - \sum^{N}_{n=1} \sum^{\rho}_{r=1} \sum^{C}_{j=1} \delta_{m^n_r, j}  \log p_j(u_r,\theta \mid I_n),
\end{equation}
for $(I_n, M_n) \in \mathbfcal{T}$, where $\delta_{m^n_r, j}$ is the Kronecker delta.

\noindent\textbf{Self-Supervised Warping:} In order to aggregate scene-level context for learning consistent semantics, we first leverage the estimated relative pose from the odometry stream to warp feature maps from the previous timestep into the current view using a predicted depth map. We then fuse the warped feature maps with the intermediate network representations of the current timestep. By incorporating feature maps from multiple views and resolutions using the representational warping concept from multi-view geometry, we enable our model to be robust to camera angle deviations, object scale, frame-level distortions and implicitly introduce feature augmentation which facilitates faster convergence. We utilize DispNet~\cite{MIFDB16} to obtain the depth map $D_t$ and fuse warped feature maps as described in~\secref{sec:awf}. We introduce the warping and fusion layers (red and orange blocks in \figref{fig:vlocnet++}) at \textit{Res4f} and \textit{Res5c} to fuse the corresponding feature maps $z_{t-1}^{4f}$ and $z_{t-1}^{5c}$ from the previous timestep into the network. As the warping is fully differentiable, our approach does not require any pre-computation for training and runs online. Moreover, our self-supervised warping adds minimal overhead as we only calculate the warping grid once at the input resolution in terms of pixels $u_{r}$ and employ average pooling to apply the grid at multiple scales for transforming the feature maps $z_{t-1}$ to its warped current view representation $\hat{z}_{t-1}$. In order to facilitate computation of gradients necessary for back-propagation, we use bilinear interpolation as a sampling mechanism for warping. Utilizing the relative pose, a depth map $D_t$ of the image, and the projection function $\pi$, we formulate the warping as
\begin{equation}
\hat{u}_r := \pi\left(T\left(\mathbf{p}_{t, t-1}\right) \pi^{-1}\left(u_{r}, D_t\left(u_{r}\right)\right)\right). \label{eq:warping}
\end{equation}
Given a previous image $I_{t-1}$ and the relative motion between the images $\mathbf{p}_{t, t-1}$, we can project each pixel $u_{r}$ from $I_{t-1}$ to $I_{t}$ as per~\eqref{eq:warping}. The warped pixel $\hat{u}_r$ is obtained using the depth information $D_t\left(u_{r}\right)$ and the relative pose $\mathbf{p}_{t, t-1}$, where the function $T\left(\mathbf{p}_{t, t-1}\right)$ denotes the homogenous transformation matrix of $\mathbf{p}_{t, t-1}$, $\pi$ denotes the projection function transforming from world to camera coordinates such that $\pi : \mathbb{R}^3 \mapsto \mathbb{R}^2$ and $\pi^{-1}$ denotes the transformation from camera to world coordinates using a depth map $D_t\left(u_{r}\right)$.
\vspace{-0.3cm}
\subsection{Deep Multitask Learning}
\label{sec:awf}

Our main motivation for jointly learning semantics, global pose regression and odometry is twofold: to enable inductive transfer by leveraging domain specific information while simultaneously exploiting complementary features, and to enable the global pose regression network to encode geometric and semantic knowledge of the environment while training. To achieve this goal, we structure our multitask framework to be interdependent on the outputs as well as intermediate representations of each of these tasks. Specifically, as shown in \figref{fig:vlocnet++}, we employ hybrid hard parameter sharing until the end of the \textit{Res3} block between the global pose regression stream and the odometry stream that both receive the image from the current timestep. This exploits the task-specific similarities among these pose regression tasks and influences the shared weights of global pose regression network to integrate motion-specific features due to inductive bias from odometry estimation, in addition to effectuating implicit attention on regions  that are more informative for relative motion estimation. 

A common practice employed for combining features from multiple layers or multiple networks is to perform concatenation of the tensors or element-wise addition/multiplication. Although this might be effective when both tensors contain sufficient relevant information, it often accumulates irrelevant feature maps and its effectiveness highly depends upon the intermediate stages of the network where the fusion is performed. One of the key components of our multitask learning framework is the proposed adaptive weighted fusion layer which learns the most favorable element-wise weighting for the fusion based on the activations in the region, followed by a non-linear feature pooling over the weighted tensors. Pooling in the feature space (as opposed to spatial pooling) is a form of coordinate-dependent transformation which yields the same number of filters as the input tensor. For ease of notation, we formulate the mathematical representation of the adaptive weighted fusion layer with respect to two activation maps $z^a$ and $z^b$ from layers $a$ and $b$, while extending the notation to multiple activation maps is straightforward. The activation maps can be from layers in the same network or from different task-specific networks. The output of the adaptive weighted fusion layer can be formulated as\looseness=-1
\begin{equation}
\hat{z}_{fuse} = \max \left( \mathbf{W} \ast \left( \left( w^a \odot z^{a} \right) \OPLUS \left( w^b \odot z^{b} \right) \right) + \mathbf{b} , 0 \right),
\end{equation}
where $w^a$ and $w^b$ are learned weightings having the same dimensions as $z^a$ and $z^b$; $\mathbf{W}$ and $\mathbf{b}$ are the parameters of the non-linear feature pooling; with $\odot$ and $\OPLUS$ representing per-channel scalar multiplication and concatenation across the channels; and $\ast$ representing the convolution operation. In other words, each channel of the activation map $z^a$ is first weighted, then linearly combined with the corresponding weighted channels of the activation map $z^b$. Non-linear feature pooling is then applied, which can be easily realized with existing layers in the form of a $1\times1$ convolution with a non-linearity such as ReLUs. As shown in \figref{fig:vlocnet++}, we incorporate the adaptive fusion layers (red blocks) at \textit{Res4c} to fuse semantic features into the global pose regression stream. In addition, we also employ them to fuse warped semantic feature maps from the previous timestep into the segmentation stream at the end of \textit{Res3} and \textit{Res4} blocks. We denote this architecture as VLocNet$++_{\text{MTL}}$ in our experiments. Moreover, in \secref{sec:mtl}, we demonstrate that over simple concatenation, our adaptive weighted fusion learns what features are relevant for both inter-task and intra-task fusion. In order to jointly learn all tasks, we minimize the loss function below:
\begin{multline}\label{eq:jointLoss}
\mathcal{L}_{multi} := \mathcal{L}_{loc} \exp(-\hat{s}_{loc}) + \hat{s}_{loc} + \mathcal{L}_{vo} \exp(-\hat{s}_{vo}) + \hat{s}_{vo} \\ + \mathcal{L}_{seg} \exp(-\hat{s}_{seg}) + \hat{s}_{seg},
\end{multline} 
where $\mathcal{L}_{loc}$ is the global pose regression loss as per~\eqref{eq:locLoss}; $\mathcal{L}_{vo}$ is the visual odometry loss from~\eqref{eq:voLoss}, and $\mathcal{L}_{seg}$ is the cross-entropy loss for semantic segmentation from ~\eqref{eq:entropyloss}. Due to the inherent nature of the diverse tasks at hand, each of the associated task-specific loss terms has a different scale. If the task-specific losses were to be naively combined, the task with the highest scale would dominate during training and there would be little if no gain for any of the other tasks. To counteract this problem, we use learnable scalar weights $\hat{s}_{loc}, \hat{s}_{vo}, \hat{s}_{seg}$ to balance the scale of each of the loss terms.
\vspace{-0.2cm}
\subsection{Datasets and Augmentation}
\label{sec:datasetsAugmentation}

Supervised learning techniques such as DCNNs require a large amount of training data with ground truth annotations which is laborious to acquire. This becomes even more critical for jointly learning multiple diverse tasks which necessitate individual task-specific labels. Although there are publicly available task-specific datasets for visual localization and semantic segmentation, to the best of our knowledge there is a lack of a large enough dataset that contains both semantic and global localization ground truth with multiple loops in the same scene. To this end, we introduce the challenging \textit{DeepLoc} dataset containing RGB-D images tagged with 6-DoF poses and pixel-level semantic labels of an outdoor urban scene that we make publicly available. In addition to our new dataset, we also benchmark the performance of our localization network (without joint semantics learning) on the challenging Microsoft 7-Scenes dataset. We chose these datasets based on the criteria of having diversity in scene structure and environment as well as the medium with which the images were captured. 

We do not perform any pose augmentations~\cite{naseer17iros, wu2017} as our initial experiments employing them did not demonstrate any improvement in performance in the aforementioned datasets. However for learning semantics, we randomly apply image augmentations including rotation, translation, scaling, skewing, cropping, flipping, contrast and brightness modulation.

\noindent\textbf{Microsoft 7-Scenes} dataset~\cite{shotton2013cvpr} is a widely used dataset for camera relocalization and tracking. It contains RGB-D image sequences tagged with 6-DoF camera poses of 7 different indoor environments. The data was captured with a Kinect camera at a resolution of $640 \times 480$ pixels and ground truth poses were generated using KinectFusion~\cite{shotton2013cvpr}. Each of the sequences contains about 500 to 1000 frames. This dataset is very challenging as it contains textureless surfaces, reflections, motion blur and perceptual aliasing due to repeating structures.

\noindent\textbf{DeepLoc}: We introduce a large-scale urban outdoor localization dataset collected around the university campus, which we make publicly available~\footnote{VLocNet++ live demo and dataset are publicly available at:\\ {\color{red}\url{http://deeploc.cs.uni-freiburg.de}}}. The dataset was collected using our robot platform equipped with a ZED stereo camera, an XSens IMU, a Trimble GPS Pathfinder Pro and several LiDARs. RGB and depth images were captured at a resolution of $1280 \times 720$ pixels, at $20\hertz$. The dataset was collected in an area spanning $110 \times 130 \meter$, that the robot traverses multiple times with different driving patterns. We use the LiDAR-based SLAM system from K{\"u}mmerle~\textit{et al.}~\cite{kummerle2015autonomous} to compute the ground truth pose labels.

Furthermore, for each image we provide pixel-level semantic segmentation annotations for ten categories: \textit{Background, Sky, Road, Sidewalk, Grass, Vegetation, Building, Poles \& Fences, Dynamic and Other}. To the best of our knowledge, this is the first publicly available dataset containing images tagged with 6-DoF poses and pixel-level semantic segmentation labels for an entire scene with multiple loops. We divide the dataset into a train and a test split such that the training set consists of seven loops with alternating driving styles amounting to 2737 images, while the test set consists of three loops with a total of 1173 images. This dataset can be very challenging for vision based applications such as global localization, camera relocalization, semantic segmentation, visual odometry and loop closure detection, as it contains substantial lighting, weather changes, motion blur and perceptual aliasing due to similar buildings and glass structures. We hope that this dataset enables future research in multitask and multimodel learning.
\vspace{-0.1cm}
\section{Experimental Evaluation}
\label{sec:results}

In order to quantify the performance of VLocNet++, we first compare our single-task models against other deep learning based methods in each corresponding task in~\secref{sec:sota}, followed by a more comprehensive comparison against the state-of-the-art in \secref{sec:benchmarking} and with multitask variants in \secref{sec:mtl}. Furthermore, we present extensive qualitative experiments and an ablation study in~\secref{sec:quant} which demonstrates the efficacy of our approach and provides insights on the representations learned by our network. For all the experiments, we train our models from random crops of the image and test on the center crop. We initialize the five residual blocks of our tasks-specific networks with weights from the ResNet-50 model trained on the ImageNet dataset and the other layers with Xavier initialization. We use the Adam solver for optimization with  $\beta_1=0.9, \beta_2=0.999$ and $\epsilon=10^{-10}$. We employ a multi-stage training procedure and first train task-specific models individually using an initial
learning rate of $\lambda_0 = 10^{-3}$ with a mini-batch size of $32$ and a dropout probability of $0.2$. Using transfer learning, we initialize the joint MTL architecture with weights from the best performing single-task models and train with a lower learning
rate of $\lambda_0 = 10^{-4}$. We use TensorFlow for the implementation and training the network on a single NVIDIA Titan X GPU takes 23 hours for the model to converge.
\vspace{-0.5cm}
\subsection{Comparison with the State-of-the-art}
\label{sec:sota}

In this section, we show empirical evaluations comparing each of the single-task models VLocNet$++_{\text{STL}}$ with other CNN-based methods for each of the corresponding tasks. 

\noindent\textbf{Evaluation of Visual Localization}: As a primary evaluation criteria, we first report results in comparison to deep learning-based approaches on both the publicly available Microsoft 7-Scenes (indoor) and DeepLoc (outdoor) datasets. We analyze the
performance in terms of the median translation and orientation errors for each scene using the original train and test splits provided by the datasets. \tabref{tab:7scenesCompGP} shows the results for the 7-Scenes dataset, for which VLocNet$++_{\text{STL}}$ achieves an overall improvement of $54.17\%$ in translation and $63.42\%$ in rotation, thereby substantially outperforming existing CNN-based approaches. The largest improvement was obtained in the perceptually hardest scenes that contain textureless regions and repeating structures such as in the stairs scene shown in \figref{fig:qualitativeLocl}(a). In this scene, we achieve an improvement of $78.35\%$ in translation and $83.34\%$ in rotation over the previous state-of-the-art. \tabref{tab:FLComp} shows the results on the DeepLoc dataset, for which we obtain almost half the error as previous methods. This demonstrates that VLocNet$++_{\text{STL}}$ performs equally well in outdoor environments where there is a significant amount of perceptual aliasing as well as in indoor textureless environments.\looseness=-1

\begin{table}
\footnotesize 
\centering
\vspace{-0.35cm}
\caption{Median localization error on the 7-Scenes dataset.}
\vspace{-0.22cm}
\label{tab:7scenesCompGP}
\begin{tabular}{@{}p{1.1cm}p{1.3cm}p{1.3cm}p{1.55cm} | p{1.9cm}@{}}
\hline\noalign{\smallskip}
Scene & PoseNet2~\cite{kendall2017geometric} & NNnet~\cite{laskar2017camera} & VLocNet~\cite{valada18icra} & VLocNet$++_{\text{STL}}$ (Ours) \\
\noalign{\smallskip}\hline\hline\noalign{\smallskip}
Chess & $0.13\meter,\, 4.48\degree$  & $0.13\meter,\, 6.46\degree$ & $0.036\meter,\, 1.71\degree$ & $\mathbf{0.023\meter ,\, 1.44\degree}$ \\ 
Fire & $0.27\meter,\, 11.3\degree$  & $0.26\meter,\, 12.72\degree$  & $0.039\meter,\, 5.34\degree$ & $\mathbf{0.018\meter,\, 1.39\degree}$ \\ 
Heads & $0.17\meter,\, 13.0\degree$  & $0.14\meter,\, 12.34\degree$ & $0.046\meter, 6.64\degree$ & $\mathbf{0.016\meter,\, 0.99\degree}$ \\ 
Office & $0.19\meter,\, 5.55\degree$  & $0.21\meter,\, 7.35\degree$  & $0.039\meter,\, 1.95\degree$ & $\mathbf{0.024\meter ,\, 1.14\degree}$ \\ 
Pumpkin & $0.26\meter,\, 4.75\degree$  & $0.24\meter,\, 6.35\degree$ & $0.037\meter,\, 2.28\degree$ & $\mathbf{0.024\meter ,\, 1.45\degree}$ \\ 
RedKitchen &  $0.23\meter,\, 5.35\degree$  & $0.24\meter,\, 8.03\degree$ & $0.039\meter,\, 2.20\degree$ & $\mathbf{0.025\meter ,\, 2.27\degree}$ \\ 
Stairs & $0.35\meter,\, 12.4\degree$  & $0.27\meter,\, 11.82\degree$ & $0.097\meter,\, 6.48\degree$ & $\mathbf{0.021\meter ,\, 1.08\degree}$ \\ 
\noalign{\smallskip}\hline\noalign{\smallskip}
Average & $0.23\meter,\, 8.12\degree$ & $0.21\meter,\, 9.30\degree$ & $0.048\meter,\, 3.80\degree$ & $\mathbf{0.022\meter ,\, 1.39\degree}$ \\ 
\noalign{\smallskip}\hline\noalign{\smallskip}
\end{tabular}
\vspace{-0.3cm}
\end{table}

\begin{table}
\footnotesize 
\centering
\caption{Median localization error on the DeepLoc dataset.}
\vspace{-0.22cm}
\label{tab:FLComp}
\begin{tabular}{@{}p{1.3cm}p{1.3cm}p{1.2cm}p{1.5cm}|p{1.9cm}@{}}
\hline\noalign{\smallskip}
PoseNet~\cite{kendall2015convolutional} & Bayesian PoseNet~\cite{kendall2015modelling} & SVS-Pose~\cite{naseer17iros} & VLocNet~\cite{valada18icra} & VLocNet$++_{\text{STL}}$ (Ours) \\
\noalign{\smallskip}\hline\hline\noalign{\smallskip}
$2.42\meter,\, 3.66\degree$  & $2.24\meter,\, 4.31\degree$ & $1.61\meter,\, 3.52\degree$ & $0.68\meter,\, 3.43\degree$  & $\mathbf{0.37\meter,\, 1.93\degree}$ \\
\noalign{\smallskip}\hline\noalign{\smallskip}
\end{tabular}
\vspace{-0.6cm}
\end{table}

\noindent\textbf{Evaluation of Visual Odometry}: We evaluate the performance of VLocNet++ for 6-DoF visual odometry estimation and show quantitative results in \tabref{tab:7scenesCompVO} for the 7-Scenes dataset and in \tabref{tab:FLCompVO} for the DeepLoc dataset. We report the average translational and rotational errors relative to the sequence length. On the 7-Scenes dataset VLocNet++  outperforms end-to-end approaches, achieving a translational error of $1.12\%$ and rotational error of $1.09\deg/\meter$. While on the outdoor DeepLoc dataset, accurately estimating ego-motion is a rather challenging task due to the more apparent motion parallax and dynamic lighting changes. Despite this fact, VLocNet++ surpasses the accuracy of competitors with a translational error of $0.12\%$ and a rotational error of $0.024\deg/\meter$.

\begin{table}
\footnotesize 
\centering
\vspace{-0.4cm}
\caption{6DoF visual odometry on the 7-Scenes dataset [$\%,\degm$].}
\vspace{-0.28cm}
\label{tab:7scenesCompVO}
\begin{tabular}{p{1.1cm}p{1cm}p{1cm}p{0.9cm}p{1cm}|p{1.2cm}}
\hline\noalign{\smallskip}
Scene & LBO \cite{NicolaiRSSw2016} & DeepVO \cite{MohantyDeepVO} & cnnBspp \cite{melekhov2017cnnBspp} & VLocNet \cite{valada18icra} & VLocNet++ (Ours) \\
\noalign{\smallskip}\hline\hline\noalign{\smallskip}
Chess & $1.69,\,1.13$ & $2.10,\,1.15$ & $1.38,\,1.12$ & $1.14,\,0.75$ & $\mathbf{0.99,\,0.66}$ \\
Fire & $3.56,\,1.42$ & $5.08,\,1.56$ & $2.08,\,1.76$ & $1.81,\,1.92$ & $\mathbf{0.99,\,0.78}$ \\
Heads & $14.43,\,2.39$ & $13.91,\,2.44$ & $3.89,\,2.70$ & $1.82,\,2.28$ & $\mathbf{0.58,\,1.59}$ \\
Office & $3.12,\,1.92$ & $4.49,\,1.74$ & $1.98,\,1.52$ & $1.71,\,1.09$ & $\mathbf{1.32,\,1.01}$ \\
Pumpkin & $3.12,\,1.60$ & $3.91,\,1.61$ & $1.29,\,1.62$ & $1.26,\,1.11$ & $\mathbf{1.16,\,0.98}$ \\
RedKitchen & $3.71,\,1.47$ & $3.98,\,1.50$ & $1.53,\,1.62$ & $1.46,\,\mathbf{1.28}$ & $\mathbf{1.26},\,1.52$ \\
Stairs & $3.64,\,2.62$ & $5.99,\,1.66$ & $2.34,\,1.86$ & $\mathbf{1.28},\,1.17$ & $1.55,\,\mathbf{1.10}$ \\
\noalign{\smallskip}\hline\noalign{\smallskip}
Average & $4.75,\,1.79$ & $5.64,\,1.67$ & $2.07,\,1.74$ & $1.51,\,1.45$ & $\mathbf{1.12,\,1.09}$ \\
\noalign{\smallskip}\hline\noalign{\smallskip}
\end{tabular}
\vspace{-0.2cm}
\end{table}

\begin{table}
\footnotesize 
\centering
\caption{6DoF visual odometry on the on the DeepLoc dataset [$\%,\degm$].}
\vspace{-0.3cm}
\label{tab:FLCompVO}
\begin{tabular}{p{1.3cm}p{1.3cm}p{1.3cm}p{1.5cm}|p{1.3cm}}
\hline\noalign{\smallskip}
LBO~\cite{NicolaiRSSw2016} & DeepVO~\cite{MohantyDeepVO} & cnnBspp~\cite{melekhov2017cnnBspp} & VLocNet~\cite{valada18icra} & VLocNet++ (Ours) \\
\noalign{\smallskip}\hline\hline\noalign{\smallskip}
$0.41,\,0.053$ & $0.33,\,0.052$ & $0.35,\,0.049$ & $0.15,\,0.040$ & $\mathbf{0.12,\,0.024}$ \\
\noalign{\smallskip}\hline\noalign{\smallskip}
\end{tabular}
\vspace{-0.6cm}
\end{table}

\begin{table*}
\footnotesize 
\centering
\caption{Comparison of semantic segmentation performance with state-of-the-art approaches on our DeepLoc dataset.}
\vspace{-0.2cm}
\label{tab:baselineSegComp}
\begin{tabular}{p{2.2cm}p{0.9cm}p{0.9cm}p{0.9cm}p{0.9cm}p{0.9cm}p{0.9cm}p{0.9cm}p{0.9cm}p{0.9cm}|p{1cm}}
\hline\noalign{\smallskip}
Approach & Sky & Road & Sidewalk & Grass & Vegetation & Building & Poles & Dynamic & Other & Mean \\
 & \crule[sky]{0.2cm}{0.2cm} & \crule[road]{0.2cm}{0.2cm} & \crule[sidewalk]{0.2cm}{0.2cm} & \crule[grass]{0.2cm}{0.2cm} & \crule[vegetation]{0.2cm}{0.2cm} & \crule[building]{0.2cm}{0.2cm} & \crule[pole]{0.2cm}{0.2cm} & \crule[dynamic]{0.2cm}{0.2cm} & \crule[other]{0.2cm}{0.2cm} & IoU \\
\noalign{\smallskip}\hline\hline\noalign{\smallskip}
FCN-8s~\cite{long2015} & $94.65$ & $98.98$ & $64.97$ & $82.14$ & $84.47$ & $87.68$ & $45.78$ & $66.39$ & $47.27$ & $69.53$ \\
SegNet~\cite{kendall2015} & $93.42$ & $98.57$ & $54.43$ & $78.79$ & $81.63$ & $84.38$ & $18.37$ & $51.57$ & $33.29$ & $66.05$ \\
UpNet~\cite{oliveira2016} & $95.07$ & $98.05$ & $63.34$ & $81.56$ & $84.79$ & $88.22$ & $31.75$ & $68.32$ & $45.21$ & $72.92$ \\
ParseNet~\cite{liu2015} & $92.85$ & $98.94$ & $62.87$ & $81.61$ & $82.74$ & $86.28$ & $27.35$ & $65.44$ & $45.12$ & $71.47$ \\
DeepLab~v2~\cite{ChenPK0Y16} & $93.39$ & $98.66$ & $76.81$ & $84.64$ & $88.54$ & $93.07$ & $20.72$ & $66.84$ & $52.70$ & $67.54$ \\
AdapNet~\cite{valada2017adapnet} & $94.65$ & $98.98$ & $64.97$ & $82.14$ & $84.48$ & $87.68$ & $45.78$ & $66.40$ & $47.27$ & $78.59$ \\
\hline\noalign{\smallskip}
VLocNet++ (ours) & $\mathbf{95.84}$  & $\mathbf{98.99}$ & $\mathbf{80.85}$ & $\mathbf{88.15}$ & $\mathbf{91.28}$ & $\mathbf{94.72}$ & $\mathbf{45.79}$ & $\mathbf{69.83}$ & $\mathbf{58.59}$ & $\mathbf{80.44}$ \\
\noalign{\smallskip}\hline\noalign{\smallskip}
\end{tabular}
\vspace{-0.6cm}
\end{table*}

\noindent\textbf{Evaluation of Semantic Segmentation}: We present comprehensive evaluations of VLocNet++ for semantic segmentation on the DeepLoc dataset and report the Intersection over Union (IoU) score for each of the individual categories as well as the mean IoU. As shown in \tabref{tab:baselineSegComp}, VLocNet++ achieves a mean IoU of $80.44\%$, consistently outperforming the baselines in all the categories. This improvement can be attributed to both the self-supervised warping as well the inductive transfer that occurs from the training signals of the localization network, as the AdapNet model which we build upon achieves a lower performance without our proposed improvements. In addition, this enables the model to converge in about 26k iterations, whereas Adapnet requires 120k iterations to converge.
\vspace{-0.2cm}
\subsection{Benchmarking on Microsoft 7-Scenes Dataset}
\label{sec:benchmarking}

\begin{figure}
\centering
\includegraphics[width=6cm]{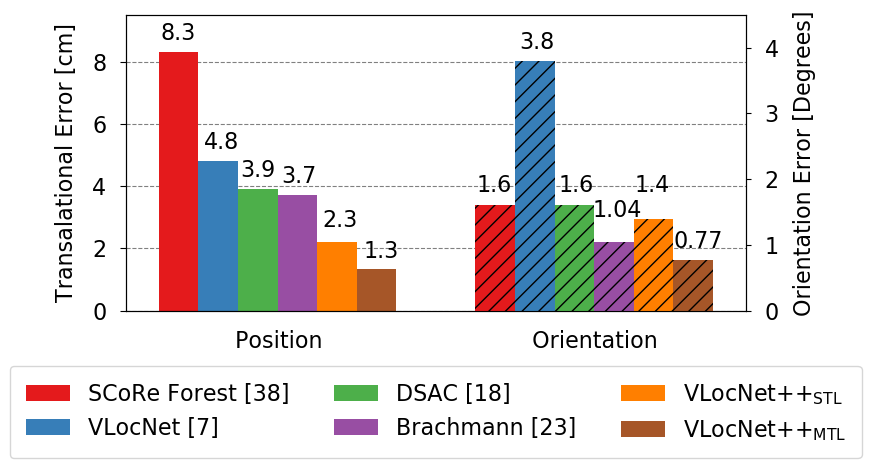}
\includegraphics[width=2.3cm]{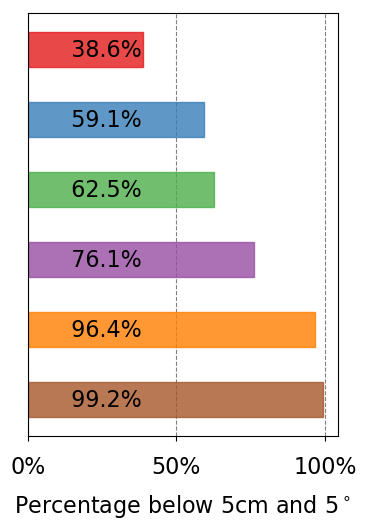}
\vspace{-0.2cm}
\caption{Benchmarking 6DoF localization on the entire 7-Scenes dataset. We compare against state-of-the-art approaches that utilize RGB or RGB-D data and even with approaches that depend on a 3D model, VLocNet++ only uses RGB images. We report the performance as median localization errors (left) and the                                                                                                                                                                percentage of test images with a pose error below $5\centi\meter$ and $5\degree$ (right).}
\label{fig:benchmarking}
\vspace{-0.4cm}
\end{figure}

We benchmark the performance of our single-task VLocNet$++_{\text{STL}}$ model and the multitask variant VLocNet$++_{\text{MTL}}$ on the Microsoft 7-Scenes dataset by comparing against both local feature-based pipelines and learning-based techniques. We present our main results in \figref{fig:benchmarking} using the median localization error metric and the percentage of poses for which the error is below $5\cm$ and $5\degree$. While VLocNet~\cite{valada18icra} was the first deep learning-based approach to yield an accuracy comparable to local feature-based pipelines achieving higher performance than SCoRe Forests~\cite{shotton2013} in terms of number of images with pose error below $5\cm$ and $5\degree$, it was recently outperformed by the approach of Brachmann~\textit{et al.}~\cite{brachmann2017} which is the current state-of-the-art.\looseness=-1

From the results presented in \figref{fig:benchmarking}, we see that our VLocNet$++_{\text{STL}}$ model achieves a localization accuracy of $96.4\%$, improving over the accuracy of Brachmann~\textit{et al.}~\cite{brachmann2017} by $20.3\%$ and by over an order of magnitude compared to the other deep learning approaches~\cite{kendall2017geometric,laskar2017camera}. Moreover, by employing our proposed multitask framework, VLocNet$++_{\text{MTL}}$ further improves on the performance and achieves an accuracy of $99.2\%$, setting the new state-of-the-art on this benchmark. Furthermore, VLocNet++ only requires $79\milli\second$ for a forward-pass on a single consumer grade GPU versus the $200\milli\second$ required by the previous state-of-the-art~\cite{brachmann2017}. It is important to note that other than VLocNet~\cite{valada18icra}, the competitors shown in \figref{fig:benchmarking} rely on a 3D scene model and hence require RGB-D data, whereas VLocNet++ only utilizes monocular images. DSAC~\cite{BrachmannKNSMGR16} and its variant~\cite{brachmann2017} that utilize only RGB images demonstrate a lower performance than the results shown in \figref{fig:benchmarking}. The improvement achieved by VLocNet++ shows that the apt combination of employing the Geometric Consistency Loss and the adaptive weighted fusion layer enables the network to efficiently leverage the motion-specific and semantic features in order to learn a geometrically consistent motion model. More extensive evaluations are shown in the supplementary material after page 8 and a live demo at {\color{red}\url{http://deeploc.cs.uni-freiburg.de}}.

\vspace{-0.2cm}
\subsection{Multitask Learning}
\label{sec:mtl}

In this section, we primarily investigate the effectiveness of employing our proposed adaptive fusion layer for encoding semantic information and aggregating motion-specific information into the global localization stream. We compare the localization accuracy of VLocNet$++_{\text{MTL}}$ which incorporates our fusion scheme against the performance of single-task models and three competitive multitask baseline approaches. A rather simple and naive approach to fuse semantic features learned by the segmentation stream into the global localization stream would be to concatenate the predicted segmentation mask as a fourth channel to the input image, which we refer to as \textit{''MTL-input-conc''}. As a second baseline, we concatenate intermediate feature maps of the segmentation stream with the corresponding intermediate feature maps in the global localization stream. We exhaustively evaluated various intermediate stages to do this fusion and in our setting we obtained the best accuracy when concatenating the feature maps of \textit{Res5c} from the segmentation with \textit{Res4f} of the global localization stream, which we denote as \textit{''MTL-mid-conc''}. For the third baseline, we share the latent space of both the networks as a variant of the approach proposed in \cite{abdulnabi2015multi}, which we denote in our experiments as \textit{''MTL-shared''}. More details on these baseline architectural topologies as well as extended ablation studies on the effect of semantic fusion and representational warping are shown in the supplementary material. \figref{fig:mtl} shows the results from this experiment on the DeepLoc dataset. VLocNet$++_{\text{MTL}}$ achieves the highest performance with an improvement of $36\%$ in translational and $53.87\%$ in rotational components of the pose, compared to the best performing MTL-input-conc baseline. While in comparison to our single-task VLocNet$++_{\text{STL}}$ model we achieve an improvement of $13.51\%$ and $23.32\%$ in the translation and rotation components respectively, demonstrating that our network is able to learn the most favorable weighting for fusion based on region activations in the feature maps. We visualize these activation maps in \figref{fig:qualitativeSeg1}.

\begin{figure}
\centering
\includegraphics[width=6cm]{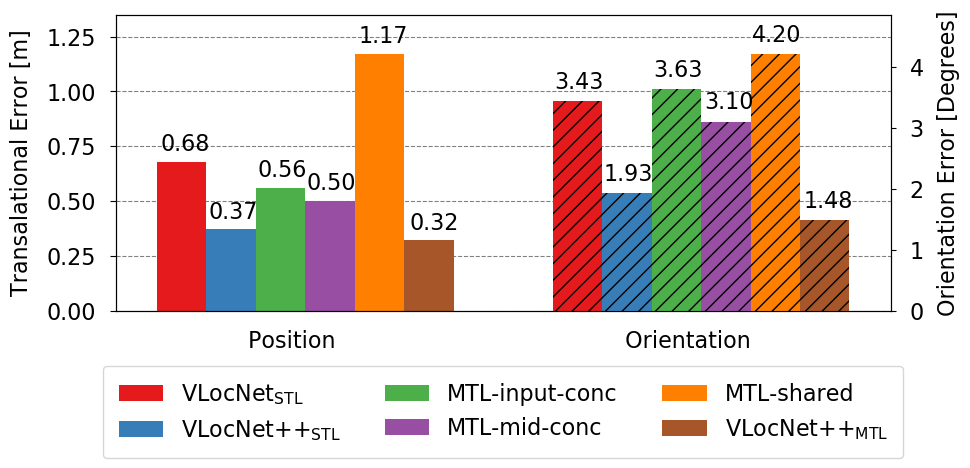}
\vspace{-0.3cm}
\caption{Localization error of various multitask models in comparison to our proposed VLocNet++ incorporating our novel adaptive weighted fusion layer for fusing semantic features into the localization stream.}
\label{fig:mtl}
\end{figure}
\vspace{-0.2cm}
\subsection{Ablation Study and Qualitative Analysis}
\label{sec:quant}

\begin{figure}
\scriptsize 
\centering 
\setlength{\tabcolsep}{0.2em} 
\begin{tabular}{p{4cm} p{4cm}}
\includegraphics[width=3.2cm]{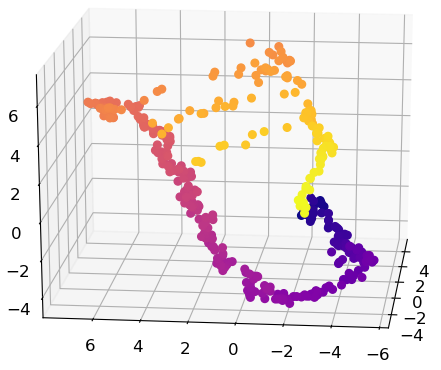} & \includegraphics[width=\linewidth]{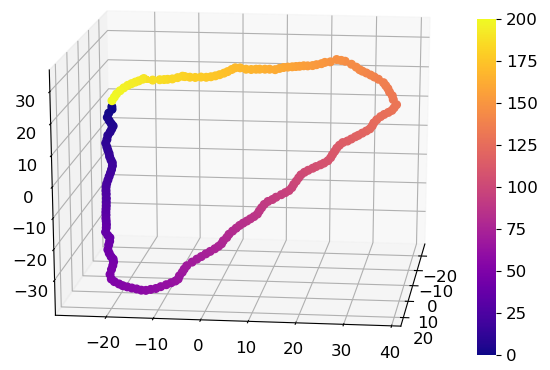} \\
\multicolumn{1}{c}{PoseNet~\cite{kendall2015convolutional}} & \multicolumn{1}{c}{VLocNet++} \\[6pt]
\end{tabular} 
\vspace{-0.3cm}
\caption{3D multi-dimensional scaling (MDS) of features from the penultimate layer of PoseNet~\cite{kendall2015convolutional} and VLocNet++ trained on the DeepLoc dataset. Inputs are images from the testing seq-01 loop and the points shown are chronologically coloured. Features learned by VLocNet++ show precise correlation with the trajectory (\figref{fig:qualitativeLocl}(b)), whereas PoseNet fails to capture the distribution especially for the poses near the glass buildings.}
\label{fig:mds}
\vspace{-0.5cm}
\end{figure}

\begin{figure}
\scriptsize 
\centering 
\setlength{\tabcolsep}{0.12cm} 
\begin{tabular}{p{2.9cm} p{5.45cm}}
\includegraphics[width=\linewidth]{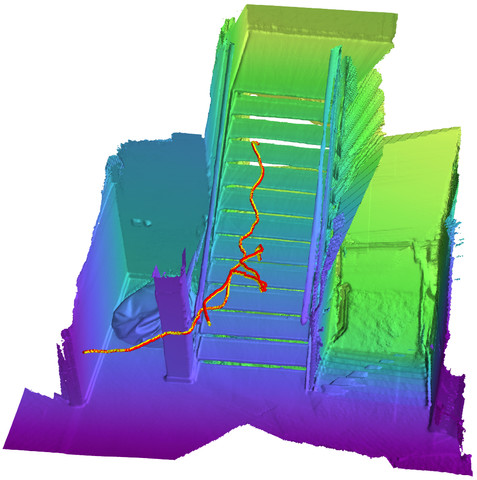} & \includegraphics[width=\linewidth]{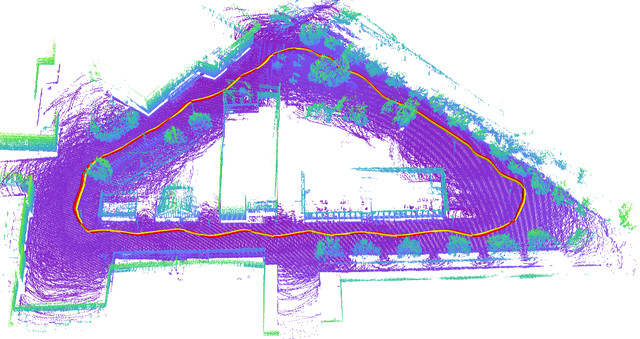} \\
\multicolumn{1}{c}{(a) Stairs} & \multicolumn{1}{c}{(b) DeepLoc} \\[6pt]
\end{tabular} 
\vspace{-0.4cm}
\caption{Qualitative localization results of one test loop depicting the estimated global pose (yellow trajectory) versus the ground truth pose (red trajectory) plotted with respect to the 3D scene model for visualization. VLocNet++ accurately estimates the pose in both indoor (a) and outdoor (b) environments while being robust to textureless regions, repetitive and reflective structures in the environment where local feature-based pipelines perform poorly.}
\label{fig:qualitativeLocl}
\end{figure}

Despite the recent surge in applying deep learning approaches to various domains, there is still a lack of fundamental knowledge regarding what kind of representations are learned by the networks, which is primarily due to their high dimensionality. To aid in this understanding, feature visualization and dimensionality reduction techniques can provide helpful insights when applied thoughtfully. Such techniques transform the data from high dimensional spaces to one of smaller dimensions by obtaining a set of principle values. For the task of localization, techniques that preserve the global geometry of the features such as Multi-Dimensional Scaling (MDS) are more meaningful to employ than approaches that find clusters and subclusters in the data such as the t-Distributed Stochastic Neighbor Embedding (t-SNE). Therefore, we apply 3D metric MDS to the features learned by the penultimate layer of our VLocNet++ model to visualize the underlying distribution. \figref{fig:mds} displays the down-projected features obtained after applying MDS for VLocNet++ and PoseNet~\cite{kendall2015convolutional} on the DeepLoc dataset. Unlike PoseNet, the features learned in VLocNet++ directly correspond to the ground truth trajectory shown in \figref{fig:qualitativeLocl}(b) (red trajectory), whereas PoseNet fails to capture the pose distribution in some areas of the dataset. Furthermore, in \figref{fig:qualitativeLocl} we show the plot of the ground truth and the estimated poses as trajectories within the 3D model of the scenes for visualization. Using our proposed VLocNet++ the estimated poses are visually indistinguishable from the ground truth demonstrating the efficacy of our approach.

\begin{figure}
\scriptsize 
\centering 
\setlength{\tabcolsep}{0.3em}
\renewcommand{\arraystretch}{1.2} 
\setlength{\fboxsep}{0pt}%
\setlength{\fboxrule}{0.5pt}%
\begin{tabular}{p{2cm} p{2cm} p{2cm} p{2cm}}
\fbox{\includegraphics[width=\linewidth]{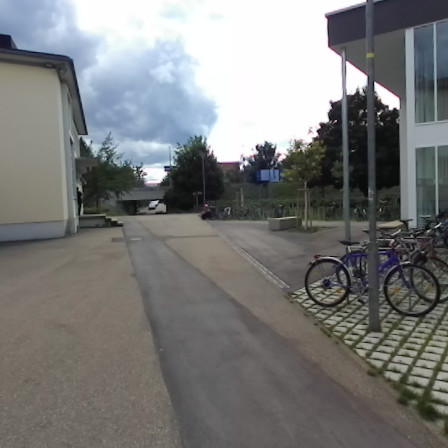}} & \includegraphics[width=\linewidth]{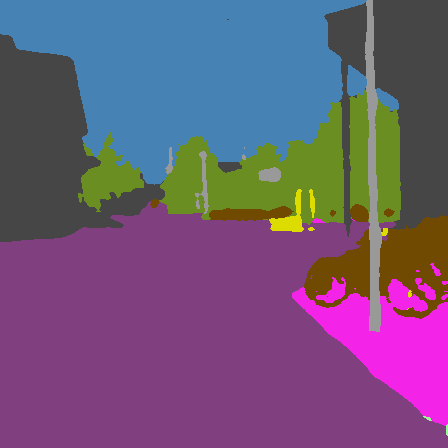} & \includegraphics[width=\linewidth, height=\linewidth]{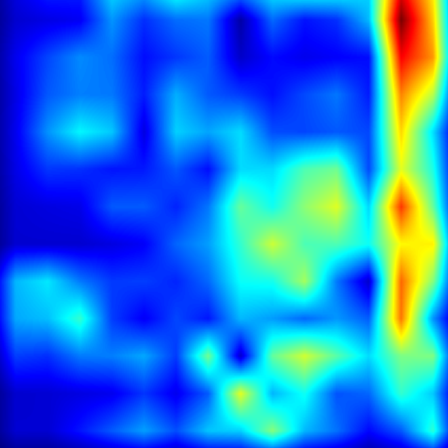} & \includegraphics[width=\linewidth, height=\linewidth]{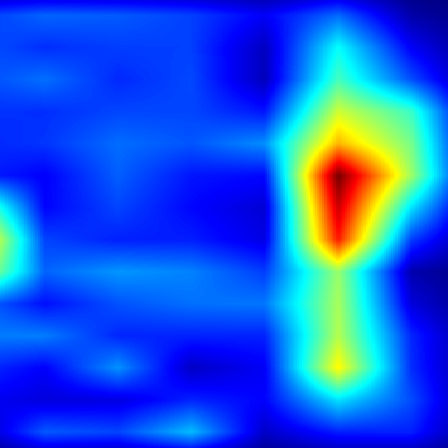} \\
\fbox{\includegraphics[width=\linewidth]{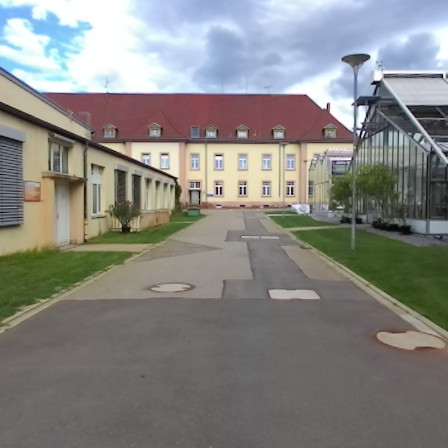}} & \includegraphics[width=\linewidth]{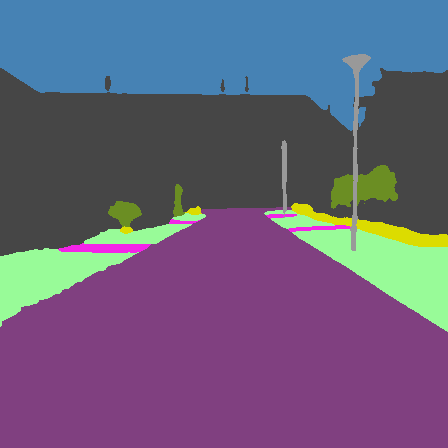} & \includegraphics[width=\linewidth, height=\linewidth]{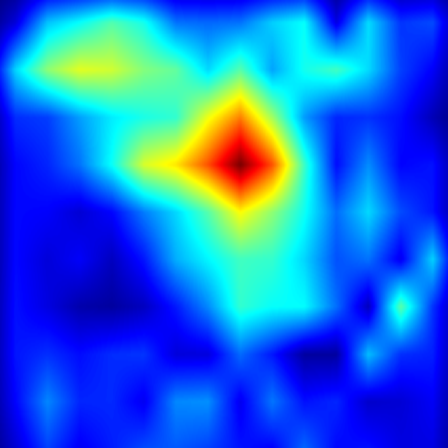} & \includegraphics[width=\linewidth]{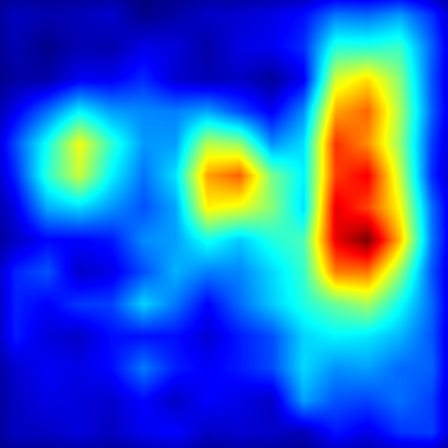} \\
(a) Input Image & (b) Semantic Output & (c) ST Activation & (d) MT Activation \\[6pt]
\end{tabular}
\vspace{-0.2cm}
\caption{Qualitative analysis of the predicted segmentation output along with a visualization of the regression activation maps~\cite{chattopadhyay2017grad} for both the single-task (ST), and multitask (MT) variant of VLocNet++ on the DeepLoc dataset.}
\label{fig:qualitativeSeg1}
\vspace{-0.5cm}
\end{figure}

In an effort to investigate the effect of incorporating semantic information on the features learned by the localization stream, we visualize the regression activation maps of the network for both the single-task and multitask variants of VLocNet++ using Grad-CAM++~\cite{chattopadhyay2017grad}. In \figref{fig:qualitativeSeg1} we show two example scenes that contain glass facades and optical glare. Despite their challenging nature, our model is able to segment both scenes with high granularity. As we compare the activation maps of our single-task and multitask models, we observe that multitask activation maps have less noisy activations focusing on multiple structures to yield an accurate pose estimate.



\section{Conclusion}
\label{sec:conclusion}

In this paper, we proposed a novel multitask learning
framework for 6-DoF visual localization, semantic segmentation and
odometry estimation, with the goal of exploiting
interdependencies within these tasks for their mutual benefit. We
presented a strategy for simultaneously encoding geometric and
structural constraints into the the pose regression network by
temporally aggregating learned motion specific information and
adaptively fusing semantic features. To this end, we proposed an
adaptive weighted fusion layer that learns the most favorable weighting for
fusion based on region activations. In addition, we proposed a
self-supervised warping technique for scene-level context
aggregation in semantic segmentation networks that improves
performance and adds minimal computational overhead while
substantially decreasing the training time. Furthermore, we introduced a large-scale outdoor localization dataset with multiple loops and pixel-level semantic ground truth for training
multitask deep networks. Comprehensive evaluations on benchmark datasets demonstrate that VLocNet++ exceeds the state-of-the-art by 67.5\% in the translational and 25.9\% in the rotational components of the pose, while being 60.5\% faster and simultaneously performing multiple tasks.



\footnotesize
\bibliographystyle{IEEEtran}
\bibliography{references}

\pagebreak

\begin{strip}
\begin{center}
\vspace{-5ex}
\textbf{\LARGE \bf
VLocNet++: Deep Multitask Learning for \\Semantic Visual Localization and Odometry} \\
\vspace{0.2cm}

\Large{\bf- Supplementary Material -}\\
\vspace{0.4cm}
\normalsize{Noha Radwan$^*$ \and Abhinav Valada$^*$ \and Wolfram Burgard}
\end{center}
\end{strip}

\setcounter{section}{0}
\setcounter{equation}{0}
\setcounter{figure}{0}
\setcounter{table}{0}
\setcounter{page}{1}
\makeatletter

%


\normalsize

This supplementary material provides details on the data collection environment and methodology, followed by in depth qualitative and quantitative experimental evaluations that were performed in addition to those reported in the main paper. We also present extensive ablation studies on the various architectural components of our network.

\section{DeepLoc Dataset}

\let\thefootnote\relax\footnote{$^*$These authors contributed equally. All authors are with the Department of Computer Science, University of Freiburg, Germany.}

\begin{figure}[!hb]
\centering
\includegraphics[width=0.4\linewidth]{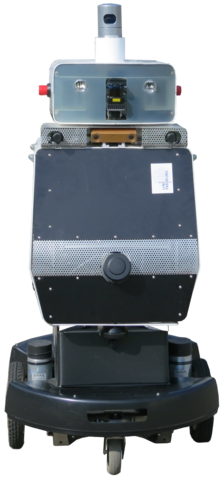}
\caption{Our robotic platform (Obelix) that was used for capturing the DeepLoc dataset. The robot is equipped with a ZED stereo camera, an XSens IMU, a Trimble GPS Pathfinder Pro and several LiDARs.}
\label{fig:obelix}
\end{figure}

\begin{figure}[!h]
\scriptsize 
\centering 
\setlength{\tabcolsep}{0.12cm} 
\begin{tabular}{p{4.2cm} p{4.2cm}}
\includegraphics[width=\linewidth]{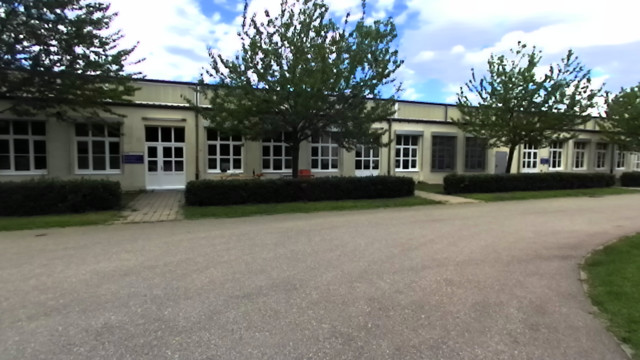} & \includegraphics[width=\linewidth]{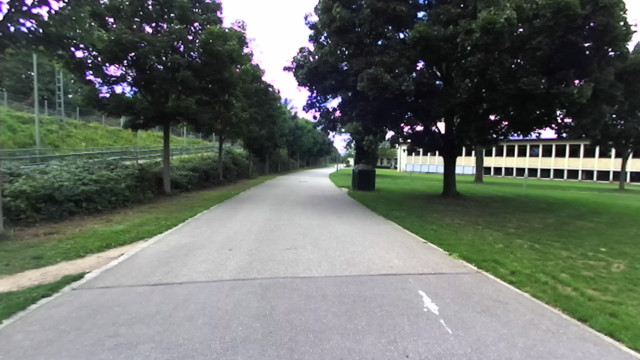}
\end{tabular} 
\caption{Example images from our DeepLoc dataset that show challenging scenarios for global pose regression and visual odometry estimation. Several sections of the traversed area contain buildings with repetitive patterns (similar to the left figure) and very few distinctive features (similar to the right figure).}
\label{fig:repBuild}
\end{figure}

\begin{figure}
\scriptsize 
\centering 
\setlength{\tabcolsep}{0.12cm} 
\begin{tabular}{p{4.2cm} p{4.2cm}}
\includegraphics[width=\linewidth]{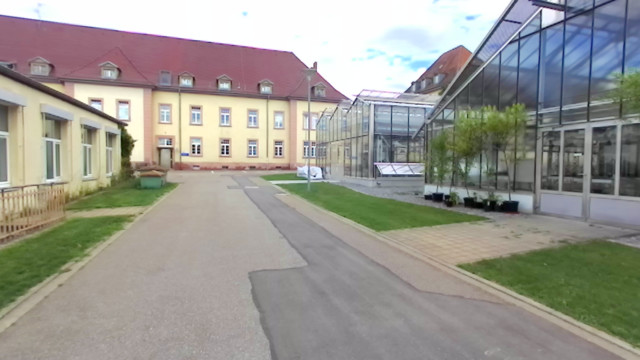} & \includegraphics[width=\linewidth]{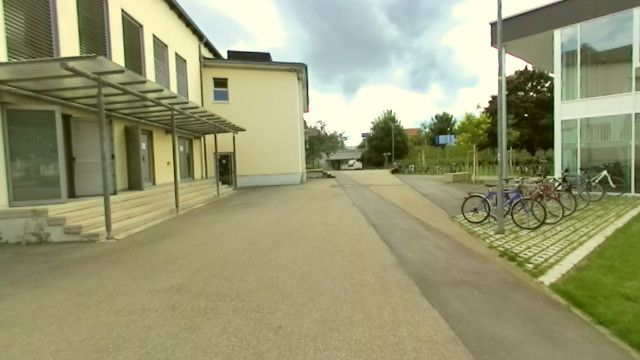}
\end{tabular} 
\caption{Examples of images from our DeepLoc dataset that are challenging for the semantic segmentation task. The traversed trajectory contained multiple structures made solely of glass (left figure), buildings with large reflective glass surfaces (right figure), as well as partially occluded structures such as bikes attached to bike-stands (right figure).}
\label{fig:diffSeg}
\end{figure}

We introduce a challenging outdoor dataset captured using our robotic platform shown in~\figref{fig:obelix}. The dataset contains stereo image pairs, depth images, pixel-level semantic labels and localization ground truth. Note that we only use monocular images for localization in this work. We make the dataset publicly available to facilitate further progress in the field of deep multi-task learning for robotics. The dataset contains multiple training and testing loops gathered around the university campus amounting to a total of ten sequences; seven of which were used for training while the rest were used for testing. Each of the sequences was collected with a different driving pattern traversing non-overlapping trajectories in order to increase the diversity of the captured data. The dataset was captured at different times of the day, therefore the images contain varying lighting conditions, reflective glare from the sun, orange dawn-sky and shadows, in addition to the motion-blur caused due to the moving robot platform. This renders the dataset challenging for a number of perception and localization tasks. Moreover, the environment that the dataset was collected in, contains structures that make semantic segmentation and localization challenging, such as buildings with similar facades, repetitive structures (see~\figref{fig:repBuild}), and translucent as well as reflective buildings made of glass. Additionally, many objects in the scene cause partial occlusion (see~\figref{fig:diffSeg}) which increase the difficulty of the semantic segmentation task.

\section{Additional Experimental Evaluations}
\label{sec:expev}

In this section, we present comprehensive additional performance evaluations of our proposed VLocNet++ in comparison to both deep learning based and local feature-based pipelines on the challenging Microsoft 7-Scenes benchmark. Additionally, we show qualitative localization results on the 7-Scenes benchmark as well as localization and segmentation results on the DeepLoc dataset in \secref{sec:qualitative}.

\subsection{Comparison with CNN-based methods}

\begin{figure*}
\footnotesize 
\centering 
\setlength{\tabcolsep}{0.2cm} 
\begin{tabular}{p{4cm} p{4cm} p{4cm} p{4cm}}
\includegraphics[width=\linewidth]{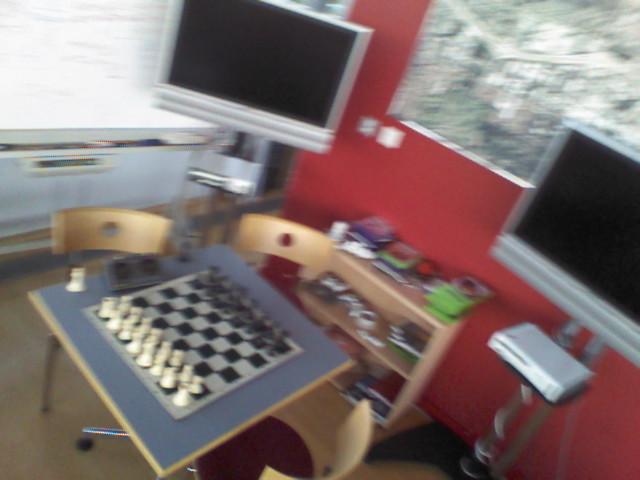} & \includegraphics[width=\linewidth]{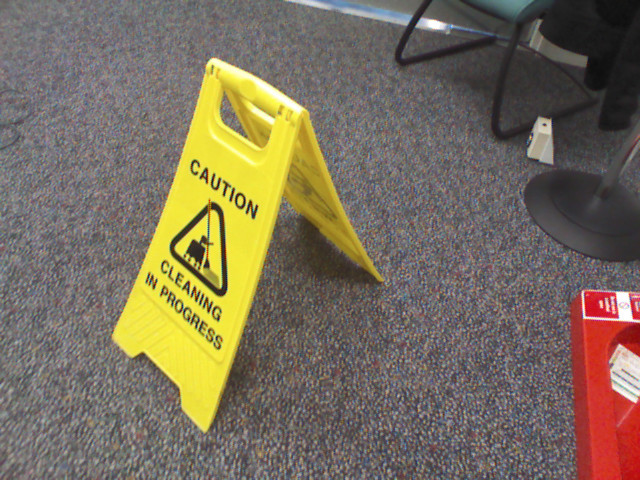} & \includegraphics[width=\linewidth]{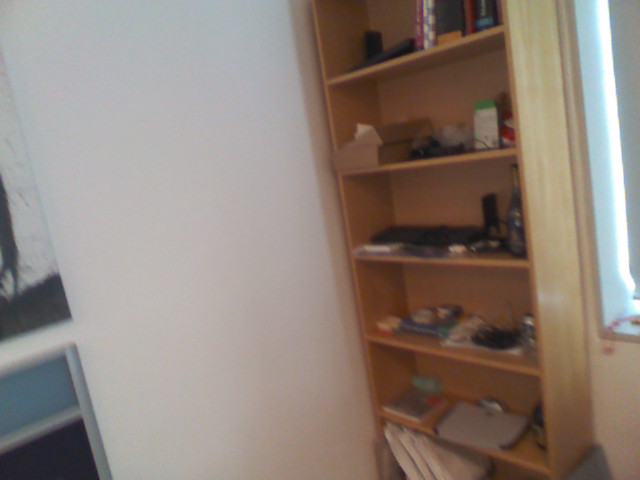} & \includegraphics[width=\linewidth]{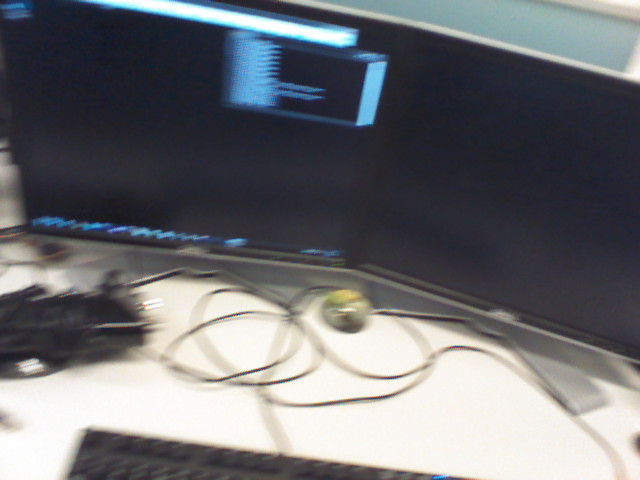} \\
\multicolumn{1}{c}{(a) Chess} & \multicolumn{1}{c}{(b) Fire} & \multicolumn{1}{c}{(c) Office} & \multicolumn{1}{c}{(d) Heads}\\
\includegraphics[width=\linewidth]{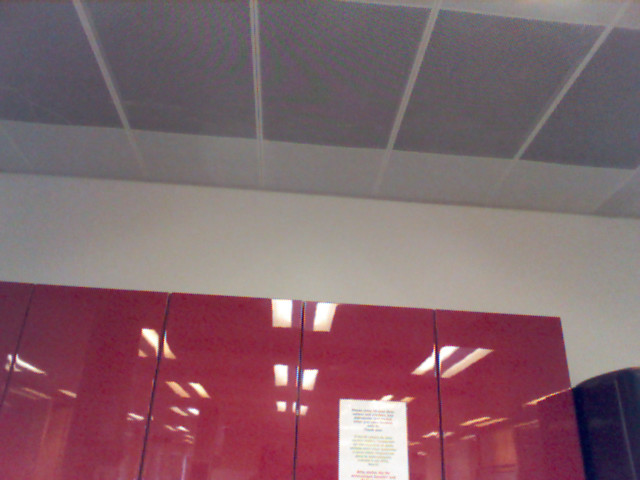} & \includegraphics[width=\linewidth]{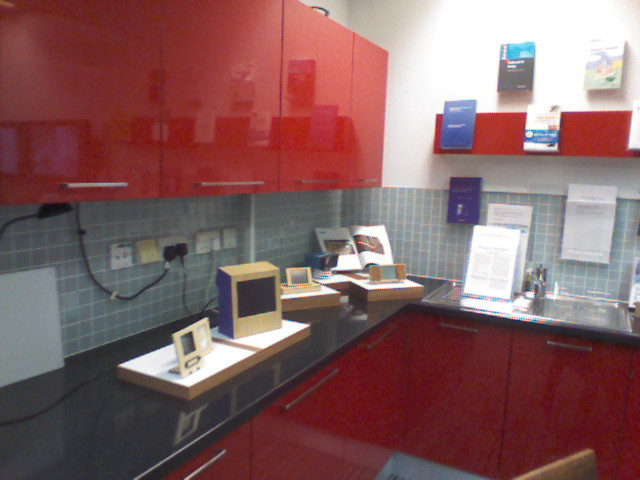} & \includegraphics[width=\linewidth]{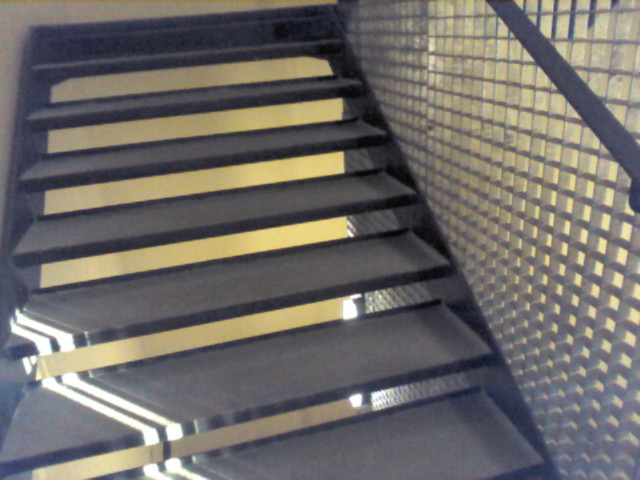} & \includegraphics[width=\linewidth]{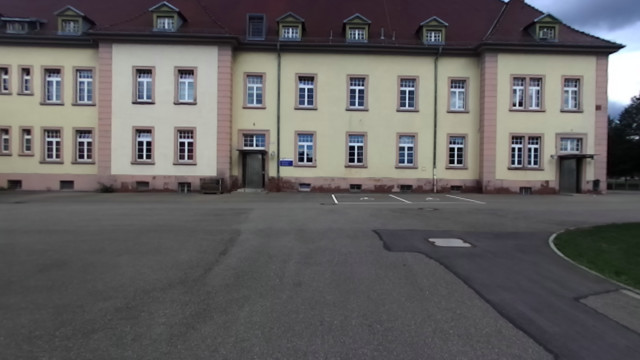} \\
\multicolumn{1}{c}{(e) Pumpkin} & \multicolumn{1}{c}{(f) Redkitchen} & \multicolumn{1}{c}{(g) Stairs} & \multicolumn{1}{c}{(h) DeepLoc} \\
\end{tabular} 
\caption{Example images from the Microsoft 7-Scenes benchmark and DeepLoc dataset that show challenging scenarios. The images exhibit significant motion blur (Chess and Heads), repetitive structures (Stairs, DeepLoc), highly reflective surfaces (Pumpkin, Redkitchen) and low-texture regions (Fire, Office). VLocNet++ is able to produce an accurate pose estimate for each of the above examples.}
\label{fig:datasetImgs}
\end{figure*}

\tabref{tab:7scenesCompGPSup} shows a comparison on the localization performance with existing deep learning based methods on the Microsoft 7-Scenes dataset. Following the standard benchmarking metrics, we report the median pose localization error. It can be seen that the proposed VLocNet++ reduces the pose error by an order of magnitude. This improvement in the localization accuracy is most apparent in scenes with repetitive structures; such as the Stairs scene (\figref{fig:datasetImgs}(g)), and scenes containing textureless and reflective surfaces such as Fire (\figref{fig:datasetImgs}(b)) and RedKitchen (\figref{fig:datasetImgs}(f)). Furthermore, by fusing the feature maps from the previous timestep using our proposed adaptive fusion layer, the network is able to better exploit the motion-specific temporal information. This leads to a significant improvement of $54.1\%$ in the translational and $63.4\%$ in the rotational components of the pose over VLocNet~\citeNew{valada18icra1}.

\begin{table*}
\footnotesize 
\centering
\caption{Median localization error of VLocNet++ with existing CNN models on the Microsoft 7-Scenes dataset.}
\label{tab:7scenesCompGPSup}
\begin{tabular}{p{1.2cm}p{1.3cm}p{1.3cm}p{1.3cm}p{1.3cm}p{1.5cm}p{1.3cm}p{1.3cm}p{1.7cm} | p{1.7cm}}
\hline\noalign{\smallskip}
Scene & PoseNet~\citeNew{kendall2015convolutional1} & Bayesian PoseNet~\citeNew{kendall2015modelling1} & LSTM-Pose~\citeNew{walch16spatialstms1} & Hourglass-Pose~\citeNew{MelekhovYKR171} & BranchNet~\citeNew{wu20171} & PoseNet2~\citeNew{kendall2017geometric1} & NNnet~\citeNew{laskar2017camera1} & VLocNet$_{\text{STL}}$~\citeNew{valada18icra1} & $\text{VLocNet++}_\text{STL}$ (Ours) \\
\noalign{\smallskip}\hline\hline\noalign{\smallskip}
Chess & $0.32\meter,\, 8.12\degree$ & $0.37\meter,\, 7.24\degree$ & $0.24\meter,\, 5.77\degree$ & $0.15\meter,\, 6.53\degree$ & $0.18\meter,\, 5.17\degree$ & $0.13\meter,\, 4.48\degree$  & $0.13\meter,\, 6.46\degree$ & $0.036\meter,\, 1.71\degree$ & $\mathbf{0.023\meter ,\, 1.44\degree}$ \\ 
Fire & $0.47\meter,\, 14.4\degree$ & $0.43\meter,\, 13.7\degree$ & $0.34\meter,\, 11.9\degree$ & $0.27\meter,\, 10.84\degree$ & $0.34\meter,\, 8.99\degree$ & $0.27\meter,\, 11.3\degree$  & $0.26\meter,\, 12.72\degree$  & $0.039\meter,\, 5.34\degree$ & $\mathbf{0.018\meter,\, 1.39\degree}$ \\ 
Heads & $0.29\meter,\, 12.0\degree$ & $0.31\meter,\, 12.0\degree$ & $0.21\meter,\, 13.7\degree$ & $0.19\meter,\, 11.63\degree$ & $0.20\meter,\, 14.15\degree$ & $0.17\meter,\, 13.0\degree$  & $0.14\meter,\, 12.34\degree$ & $0.046\meter, 6.64\degree$ & $\mathbf{0.016\meter,\, 0.99\degree}$ \\ 
Office & $0.48\meter,\, 7.68\degree$ & $0.48\meter,\, 8.04\degree$ & $0.30\meter,\, 8.08\degree$ & $0.21\meter,\, 8.48\degree$ & $0.30\meter,\, 7.05\degree$ & $0.19\meter,\, 5.55\degree$  & $0.21\meter,\, 7.35\degree$  & $0.039\meter,\, 1.95\degree$ & $\mathbf{0.024\meter ,\, 1.14\degree}$ \\ 
Pumpkin & $0.47\meter,\, 8.42\degree$ & $0.61\meter,\, 7.08\degree$ & $0.33\meter,\, 7.00\degree$ &  $0.25\meter,\, 7.01\degree$ & $0.27\meter,\, 5.10\degree$ & $0.26\meter,\, 4.75\degree$  & $0.24\meter,\, 6.35\degree$ & $0.037\meter,\, 2.28\degree$ & $\mathbf{0.024\meter ,\, 1.45\degree}$ \\ 
RedKitchen & $0.59\meter,\, 8.64\degree$ & $0.58\meter,\, 7.54\degree$ & $0.37\meter,\, 8.83\degree$ & $0.27\meter,\, 10.15\degree$ & $0.33\meter,\, 7.40\degree$ &  $0.23\meter,\, 5.35\degree$  & $0.24\meter,\, 8.03\degree$ & $0.039\meter,\, 2.20\degree$ & $\mathbf{0.025\meter ,\, 2.27\degree}$ \\ 
Stairs & $0.47\meter,\, 13.8\degree$ & $0.48\meter,\, 13.1\degree$ & $0.40\meter,\, 13.7\degree$ & $0.29\meter,\, 12.46\degree$ & $0.38\meter,\, 10.26\degree$ & $0.35\meter,\, 12.4\degree$  & $0.27\meter,\, 11.82\degree$ & $0.097\meter,\, 6.48\degree$ & $\mathbf{0.021\meter ,\, 1.08\degree}$ \\ 
\noalign{\smallskip}\hline\noalign{\smallskip}
Average & $0.44\meter,\, 10.4\degree$ & $0.47\meter,\, 9.81\degree$ & $0.31\meter,\, 9.85\degree$ & $0.23\meter,\, 9.53\degree$ & $0.29\meter,\, 8.30\degree$ & $0.23\meter,\, 8.12\degree$ & $0.21\meter,\, 9.30\degree$ & $0.048\meter,\, 3.80\degree$ & $\mathbf{0.022\meter ,\, 1.39\degree}$ \\ 
\noalign{\smallskip}\hline\noalign{\smallskip}
\end{tabular}
\end{table*}

\begin{table*}
\footnotesize 
\centering
\caption{Benchmarking the median localization error of VLocNet++ on the Microsoft 7-Scenes dataset.}
\label{tab:7scenesCompMult}
\begin{tabular}{p{1.2cm}p{1.3cm}p{1.35cm}p{1.35cm}p{1.35cm}p{1.35cm}p{1.35cm}p{1.3cm} p{1.7cm} | p{1.7cm}}
\hline\noalign{\smallskip}
Scene & Active & SCoRe & $\text{VLocNet}_\text{STL}$ & $\text{VLocNet}_\text{MTL}$ & DSAC~\citeNew{BrachmannKNSMGR161} & Brachmann & Brachmann & $\text{VLocNet++}_\text{STL}$ & $\text{VLocNet++}_\text{MTL}$ \\
 & Search~\citeNew{sattler2017efficient1} & Forest~\citeNew{shotton20131} & \citeNew{valada18icra1} & \citeNew{valada18icra1} & & \textit{et al.}~\citeNew{brachmann20171} w/o 3D & \textit{et al.}~\citeNew{brachmann20171} w/ 3D & (Ours) & (Ours) \\
\noalign{\smallskip}\hline\hline\noalign{\smallskip}
Chess & $0.04\meter,\, 1.96\degree$ & $0.03\meter,\, \mathbf{0.66}\degree$ & $0.03\meter,\, 1.71\degree$ & $0.03\meter,\,1.69\degree$ & $0.02\meter,\, 1.20\degree$ & $0.02\meter,\, 0.70\degree $ & $0.02\meter,\, 0.50\degree$  & $0.023\meter ,\, 1.44\degree$ & $\mathbf{0.018}\meter,\, 1.17\degree$\\ 
Fire & $0.03\meter,\, 1.53\degree$ &$0.05\meter,\, 1.50\degree$ & $0.04\meter,\, 5.34\degree$ & $0.04\meter ,\, 4.86\degree$& $0.04\meter,\, 1.50\degree$& $0.04\meter,\, 1.20\degree $ & $0.02\meter,\, 0.80\degree$ & $0.018\meter,\, 1.39\degree$ & $\mathbf{0.009\meter,\,0.61\degree}$\\ 
Heads & $0.02\meter,\, 1.45\degree$ & $0.06\meter,\, 5.50\degree$ & $0.04\meter,\, 6.64\degree$ & $0.05\meter,\, 4.99\degree$ & $0.03\meter,\, 2.70\degree$ & $0.24\meter,\, 10.00\degree$ & $0.01\meter,\, 0.80\degree$ & $0.016\meter,\, 0.99\degree$ & $\mathbf{0.008\meter,\, 0.60\degree}$\\ 
Office & $0.09\meter,\, 3.61\degree$ & $0.04\meter,\, 0.78\degree$ & $0.04\meter,\, 1.95\degree$ & $0.03\meter ,\, 1.51\degree$ & $0.04\meter,\, 1.60\degree$ & $0.03\meter,\, 0.80\degree$ &  $0.03\meter,\, \mathbf{0.70\degree}$ & $0.024\meter ,\, 1.14\degree$ & $\mathbf{0.016\meter},\,0.78\degree$\\ 
Pumpkin & $0.08\meter,\, 3.10\degree$ & $0.04\meter,\, \mathbf{0.68\degree}$ & $0.04\meter,\, 2.28\degree$ & $0.04\meter,\, 1.92\degree$ & $0.05\meter,\, 2.00\degree$ & $0.04\meter,\, 1.10\degree$ & $ 0.04\meter,\, 1.00\degree$ & $0.024\meter ,\, 1.45\degree$& $\mathbf{0.009\meter},\,	0.82\degree$\\  
RedKitchen & $0.07\meter ,\, 3.37\degree$ & $0.04\meter,\, \mathbf{0.76\degree}$ & $0.04\meter,\, 2.20\degree$ & $0.03\meter,\, 1.72\degree$ & $0.05\meter ,\, 2.00\degree$ & $0.05\meter,\, 1.30\degree$ & $0.04\meter,\, 1.00\degree$ &$0.025\meter ,\, 2.27\degree$ &$\mathbf{0.017\meter},\,0.93\degree$\\  
Stairs & $0.03\meter,\,2.22\degree$ & $0.32\meter,\, 1.32\degree$ & $0.10\meter,\, 6.48\degree$ & $0.07\meter,\, 4.96\degree$ & $1.17\meter ,\, 33.1\degree$ & $0.27\meter,\, 5.40\degree $ & $0.10\meter,\, 2.50\degree$ &$0.021\meter ,\, 1.08\degree$ &$\mathbf{0.010\meter,\,0.48\degree}$\\ 
\noalign{\smallskip}\hline\noalign{\smallskip}
Average & $0.05\meter,\,2.46\degree$ &$0.08\meter,\, 1.60\degree$ & $0.048\meter,\, 3.80\degree$ & $0.04\meter,\, 3.09\degree$ & $0.20\meter,\,6.30\degree$ & $0.099\meter,\,2.92\degree$ & $0.04\meter,\,1.04\degree$ & $0.022\meter ,\, 1.39\degree$ &$\mathbf{0.013\meter,\,	0.77\degree}$\\  
\noalign{\smallskip}\hline\noalign{\smallskip}
\end{tabular}
\end{table*}

\begin{table*}
\footnotesize 
\centering
\caption{Comparison with the state-of-the-art on the Microsoft 7-Scenes dataset.}
\label{tab:brachComp}
\begin{tabular}{p{2.8cm}p{2cm}p{1.5cm}p{1.8cm}p{1.7cm}p{1.7cm}}
\noalign{\smallskip}\hline\noalign{\smallskip}
Method & Input & Median Translational Error & Median Rotational Error & Pose Accuracy & Run-time\\
\noalign{\smallskip}\hline\hline\noalign{\smallskip}
Brachmann~\textit{et al.}~\citeNew{brachmann20171} & w/ 3D & $0.04\meter$ & $1.04\degree$ & $76.1\%$ & $200\milli\second$ \\
$\text{VLocNet++}_\text{MTL}$ (Ours) & Monocular & $\mathbf{0.013\meter}$ & $\mathbf{0.77\degree}$ & $\mathbf{99.2\%}$ & $\mathbf{79\milli\second}$\\
\noalign{\smallskip}\hline\noalign{\smallskip}
\end{tabular}
\end{table*}

\subsection{Benchmarking Against State-of-the-art}

In order to gain further insights on the performance of the different variants of VLocNet++ in comparison to state-of-the-art local feature-based localization methods, we report the median localization pose error on the Microsoft 7-Scenes dataset in~\tabref{tab:7scenesCompMult}. Incorporating the proposed adaptive weighted fusion approach for previous pose fusion reduces the localization error by approximately $50\%$ in comparison to VLocNet. More importantly, by jointly learning to regress the relative motion (odometry) in addition to global localization, the network is able to efficiently incorporate motion-specific features that are necessary for accurate pose predictions. This enables our network to achieve sub-centimeter and sub-degree accuracy for the majority of the scenes. Furthermore, unlike local feature-based approaches, our proposed VLocNet++ is able to accurately estimate the global pose in environments containing repetitive and textureless structures. \tabref{tab:brachComp} compares the performance of our proposed $\text{VLocNet++}_\text{MTL}$ with the approach of Brachmann~\textit{et al.}~\citeNew{brachmann20171} in terms of the median localization error, percentage of the poses with localization error below $5\centi\meter$ and $5\degree$ (denoted by pose accuracy) and run-time. The approach of Brachmann~\textit{et al.}~\citeNew{brachmann20171} is currently the state-of-the-art on this benchmark. It can be seen from \tabref{tab:brachComp} that our proposed approach exceeds the state-of-the-art by $67.5\%$ in the translational and $25.9\%$ in the rotational components of the pose. Furthermore, unlike the approach of Brachmann~\textit{et al.}~\citeNew{brachmann20171}, our proposed approach does not require a 3D model of the scene which facilitates ease of deployment, in addition to occupying less storage space for the model. Moreover, the run-time of VLocNet++ is $60.5\%$ faster (run on a single consumer grade GPU) than that of Brachmann \textit{et al.}~\citeNew{brachmann20171}. This renders our method well suited for real-time deployment in an online manner, as well as in resource restricted platforms. 

\begin{figure*}
\footnotesize 
\centering 
\setlength{\tabcolsep}{0.5cm} 
\begin{tabular}{p{5cm} p{5cm} p{5cm}}
\includegraphics[width=\linewidth]{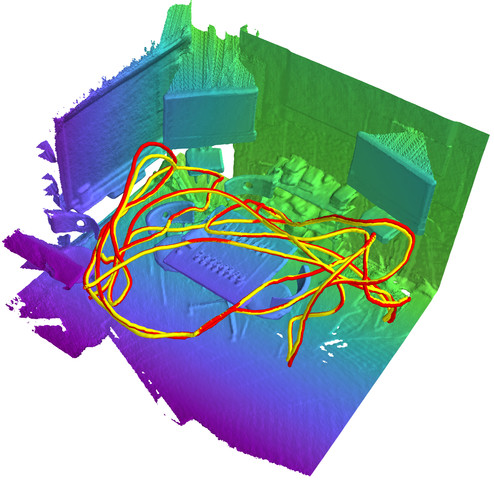} & \includegraphics[width=\linewidth]{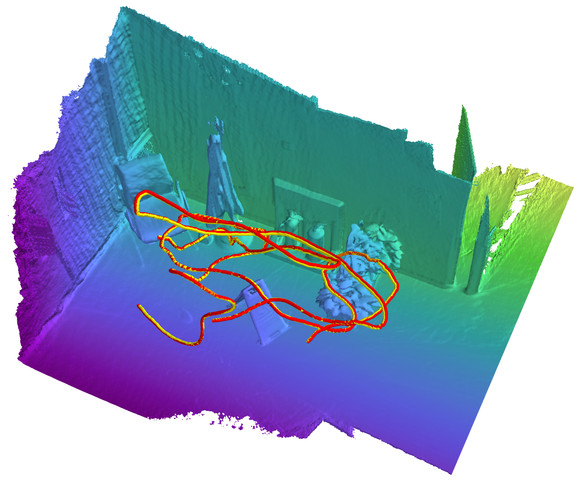} & \includegraphics[width=\linewidth]{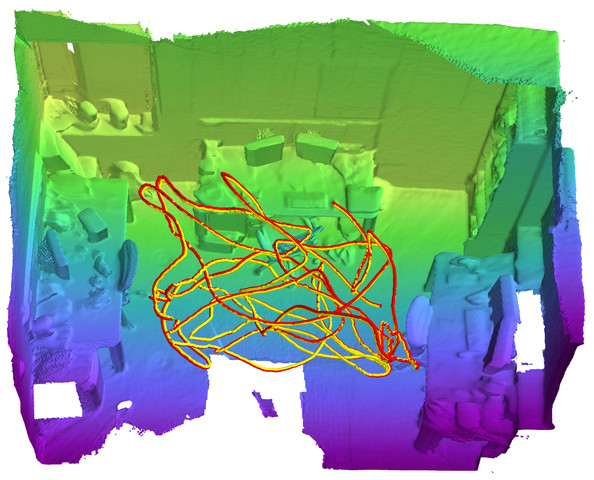} \\
\multicolumn{1}{c}{(a) Chess} & \multicolumn{1}{c}{(b) Fire} & \multicolumn{1}{c}{(c) Office} \\

\includegraphics[width=\linewidth]{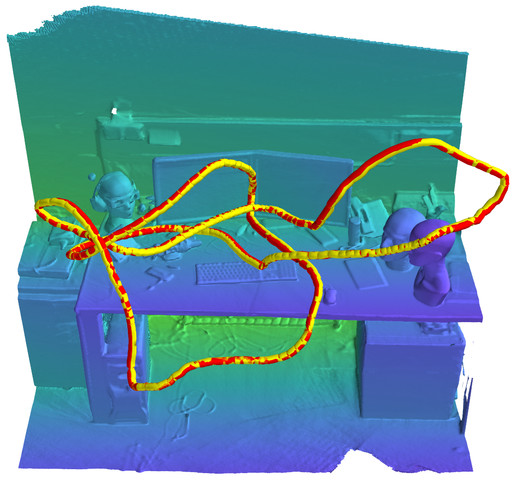} & \includegraphics[width=\linewidth]{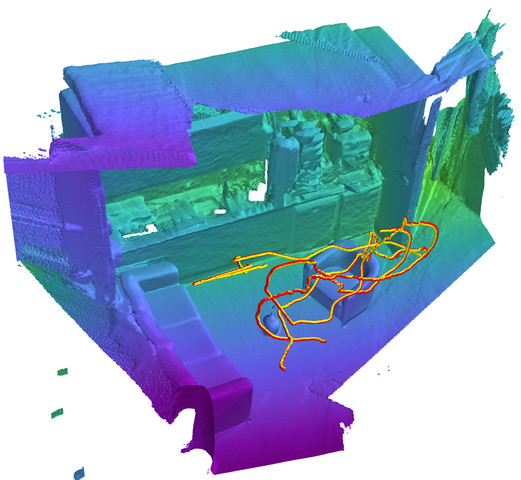} & \includegraphics[width=\linewidth]{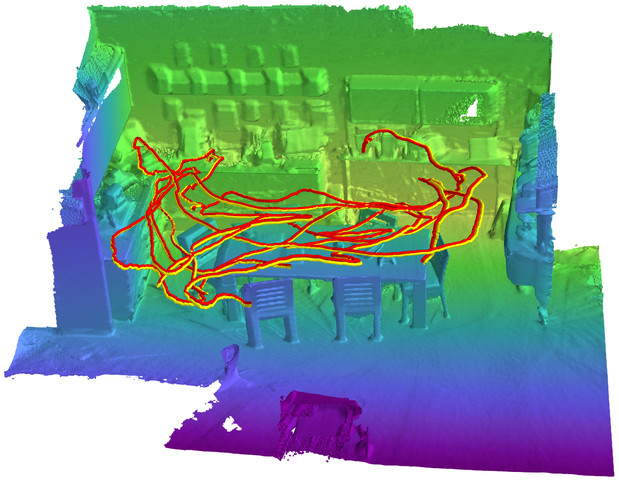} \\
\multicolumn{1}{c}{(d) Heads} & \multicolumn{1}{c}{(e) Pumpkin} & \multicolumn{1}{c}{(f) Redkitchen} \\
 \includegraphics[width=\linewidth]{stairs.jpg} & \multicolumn{2}{c}{\includegraphics[width=0.5\linewidth]{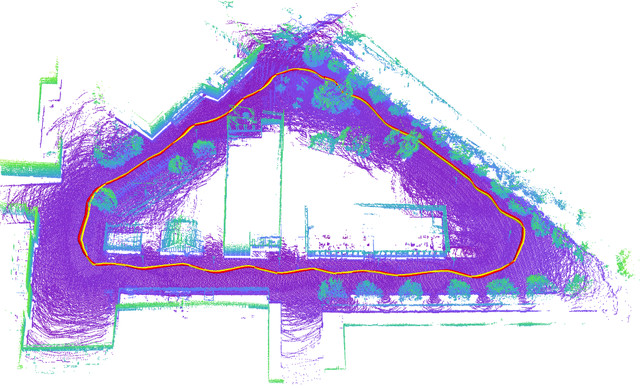}}\\
\multicolumn{1}{c}{(g) Stairs} & \multicolumn{2}{c}{(h) DeepLoc}\\
\end{tabular} 
\caption{Qualitative localization results depicting the predicted global pose (yellow trajectory) versus the ground truth pose (red trajectory) plotted with respect to the 3D scene model for visualization. VLocNet++ accurately estimates the global pose in both indoor (a, b, c, d, e, f, g) and outdoor (h) environments while being robust to textureless regions, repetitive as well as reflective structures in the environment where local feature-based pipelines perform poorly. Note that we only show the trajectory plotted with respect to the 3D scene model for visualization, our approach does not rely on a 3D model for localization. We show the second testing loop for the DeepLoc dataset, as visualizing all the testing loops in one scene creates a intertwined output that is visually hard to analyze.}
\label{fig:qualitativeLoc}
\end{figure*}

\begin{figure*}
\footnotesize 
\centering 
\setlength{\tabcolsep}{0.2cm} 
\begin{tabular}{p{0.1cm}p{4cm} p{4cm} p{4cm} p{4cm}}
\rotatebox[origin=c]{90}{Input Image} & \raisebox{-0.5\height}{\includegraphics[width=\linewidth]{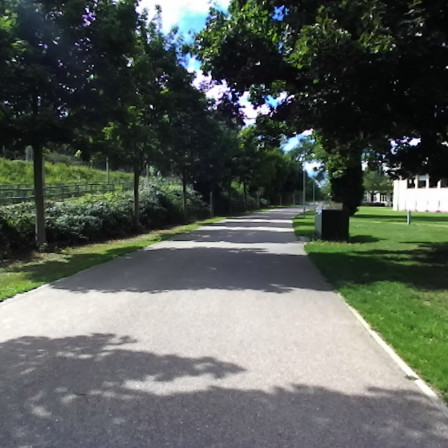}} & \raisebox{-0.5\height}{\includegraphics[width=\linewidth]{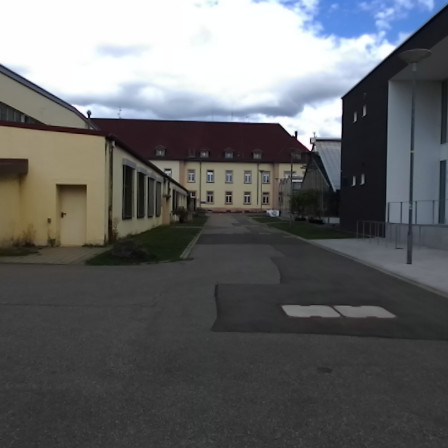}} & \raisebox{-0.5\height}{\includegraphics[width=\linewidth]{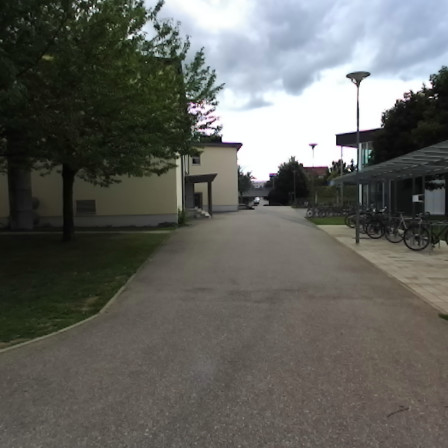}} & \raisebox{-0.5\height}{\includegraphics[width=\linewidth]{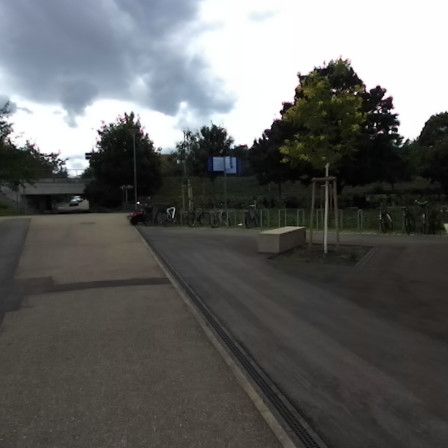}} \\
\\
\rotatebox[origin=c]{90}{Adapnet Output~\citeNew{valada2017adapnet1}} & \raisebox{-0.5\height}{\includegraphics[width=\linewidth]{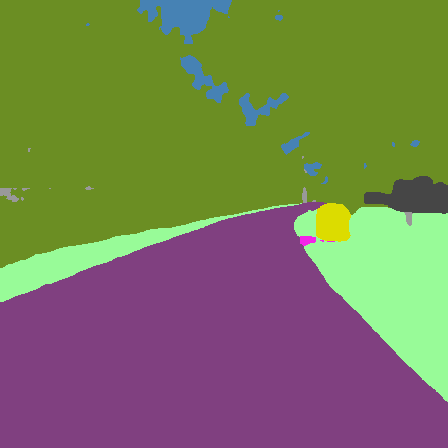}} & \raisebox{-0.5\height}{\includegraphics[width=\linewidth]{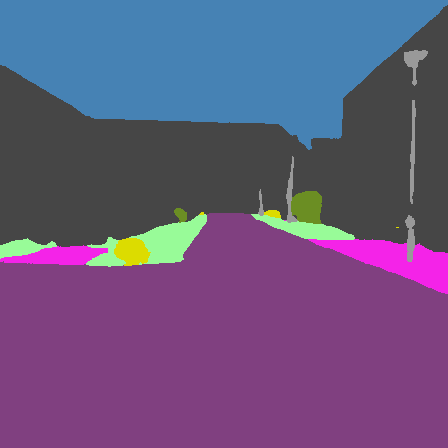}} & \raisebox{-0.5\height}{\includegraphics[width=\linewidth]{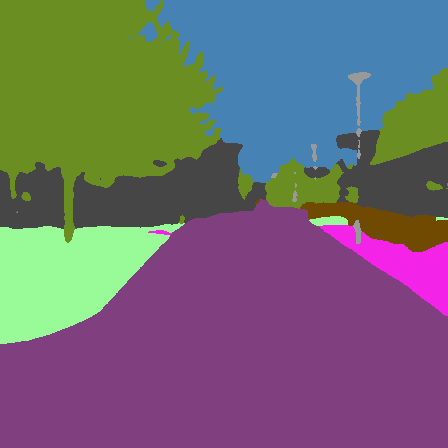}} & \raisebox{-0.5\height}{\includegraphics[width=\linewidth]{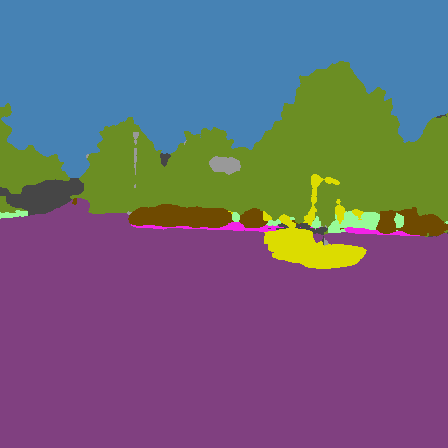}}\\
\\
\rotatebox[origin=c]{90}{VLocNet++ Output} & \raisebox{-0.5\height}{\includegraphics[width=\linewidth]{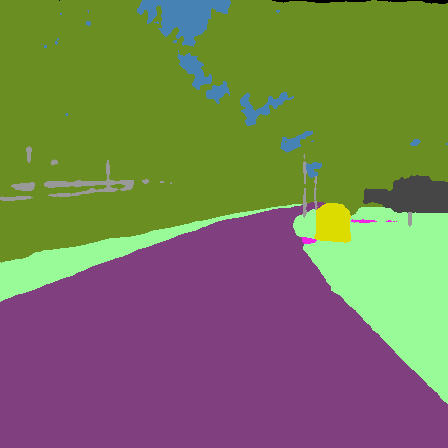}} & \raisebox{-0.5\height}{\includegraphics[width=\linewidth]{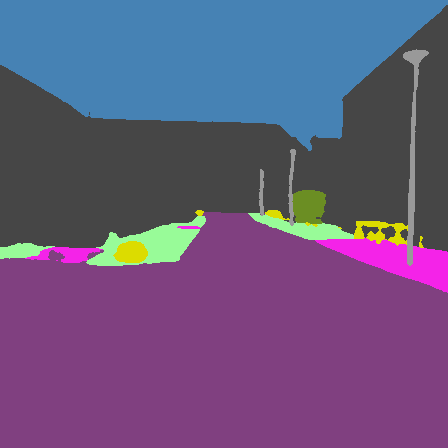}} & \raisebox{-0.5\height}{\includegraphics[width=\linewidth]{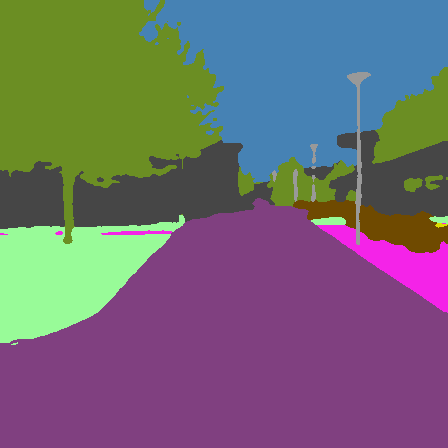}} & \raisebox{-0.5\height}{\includegraphics[width=\linewidth]{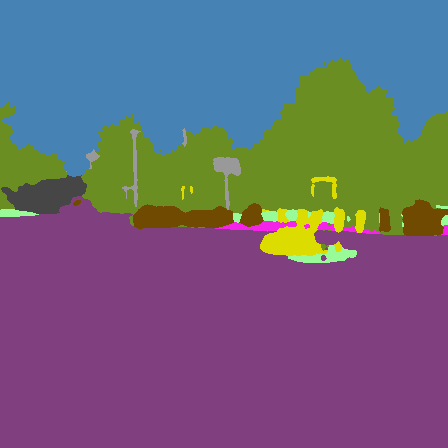}}\\
& \multicolumn{1}{c}{(a)} & \multicolumn{1}{c}{(b)} & \multicolumn{1}{c}{(c)}  & \multicolumn{1}{c}{(d)}\\
\end{tabular} 
\caption{Qualitative comparison of semantic segmentation results obtained using Adapnet~[13] versus VLocNet++ on the DeepLoc dataset. The semantic categories are color coded as follows: \crule[sky]{0.2cm}{0.2cm}~Sky, \crule[road]{0.2cm}{0.2cm}~Road, \crule[sidewalk]{0.2cm}{0.2cm}~Sidewalk, \crule[grass]{0.2cm}{0.2cm}~Grass, \crule[vegetation]{0.2cm}{0.2cm}~Vegetation, \crule[building]{0.2cm}{0.2cm}~Building, \crule[pole]{0.2cm}{0.2cm}~Poles, \crule[dynamic]{0.2cm}{0.2cm}~Dynamic and \crule[other]{0.2cm}{0.2cm}~Other. VLocNet++ accurately segments thin structures such as poles, bike stands and fences (a, b, c, d), precisely captures the boundaries between grass and vegetation (a), and segments distant thin sidewalk paths that are all not accurately segmented in the outputs obtained from AdapNet.} 
\label{fig:comparisonSeg}
\end{figure*}

\subsection{Qualitative Analysis}
\label{sec:qualitative}

The experiments presented thus far have demonstrated the capability of VLocNet++ in terms of the performance metrics. In order to analyze the localization performance qualitatively with respect to the scene, we present visualizations that show the pose estimate obtained from VLocNet++ in comparison to the ground truth on both the indoor Microsoft 7-Scenes dataset and the outdoor DeepLoc dataset. Results from this experiment are shown in \figref{fig:qualitativeLoc}. The predicted poses are shown as a yellow trajectory and the ground truth poses are shown as a red trajectory. Note that the 3D model is only shown for visualization purposes and our approach does not utilize it for localization as it only operates on monocular images. We provide visualizations of the 3D scene models along with the predicted and ground truth poses at {\color{red}\url{http://deeploc.cs.uni-freiburg.de}}.

In~\figref{fig:datasetImgs}, we present example images from both the Microsoft 7-Scenes and DeepLoc datasets. The images are representative of the challenges encountered in each scene. The images show scenarios that are challenging for global pose regression and visual odometry estimation such as substantial blur due to camera motion (\figref{fig:datasetImgs}(a, d)), perceptual aliasing due to repeating structures (\figref{fig:datasetImgs}(g, h)), textureless regions (\figref{fig:datasetImgs} (b, c)) and highly reflective surfaces (\figref{fig:datasetImgs}(e, f)). Despite these challenges, the predicted poses by VLocNet++ are well aligned with their ground truth counterparts as observed in~\figref{fig:qualitativeLoc}. By utilizing the adaptive weighted fusion and the hybrid parameter sharing between the global pose regression and visual odometry estimation streams, VLocNet++ is able to accurately capture motion-specific features without requiring a 3D model of the environment.

\begin{figure*}
\footnotesize 
\centering 
\setlength{\tabcolsep}{0.2cm} 
\begin{tabular}{p{0.1cm}p{4cm} p{4cm} p{4cm} p{4cm}}
\rotatebox[origin=c]{90}{Input Image} & \raisebox{-0.5\height}{\includegraphics[width=\linewidth]{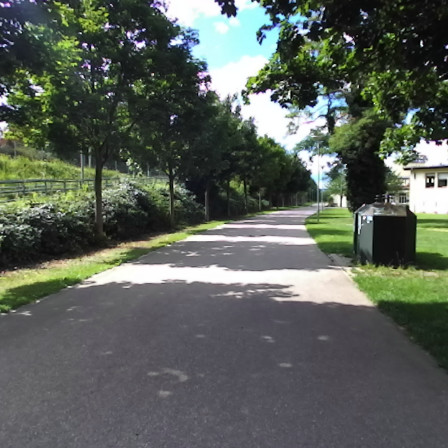}} & \raisebox{-0.5\height}{\includegraphics[width=\linewidth]{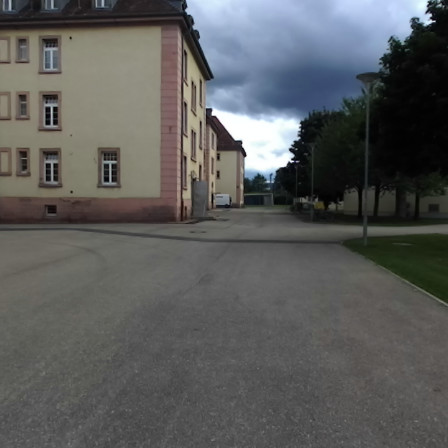}} & \raisebox{-0.5\height}{\includegraphics[width=\linewidth]{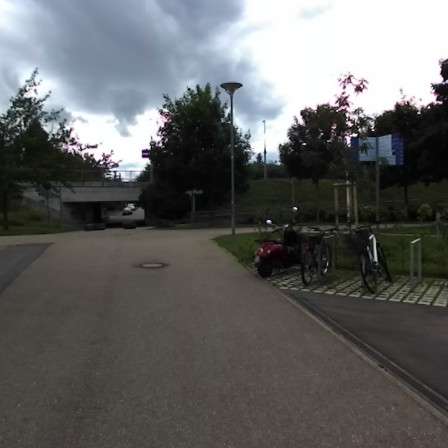}} & \raisebox{-0.5\height}{\includegraphics[width=\linewidth]{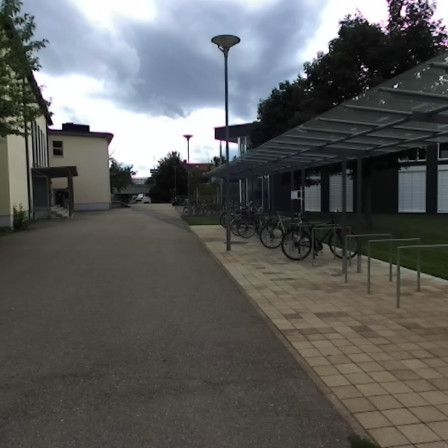}} \\
\\
\rotatebox[origin=c]{90}{Segmented Output} & \raisebox{-0.5\height}{\includegraphics[width=\linewidth]{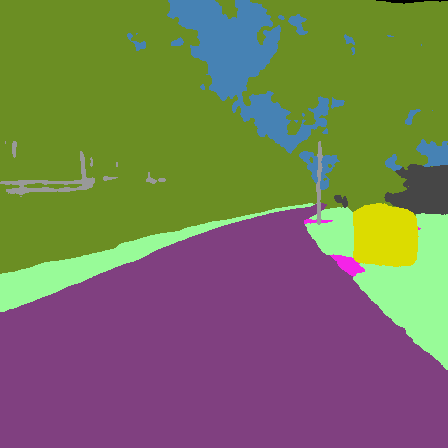}} & \raisebox{-0.5\height}{\includegraphics[width=\linewidth]{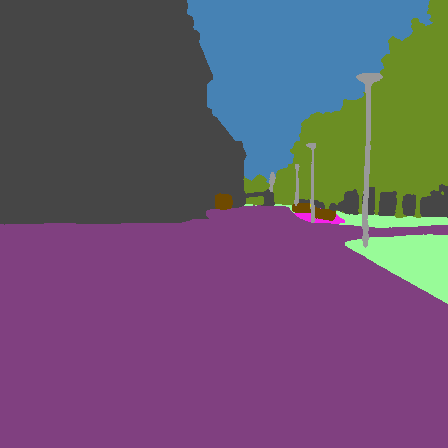}} & \raisebox{-0.5\height}{\includegraphics[width=\linewidth]{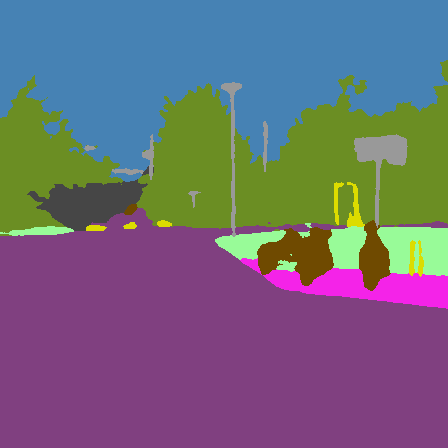}} & \raisebox{-0.5\height}{\includegraphics[width=\linewidth]{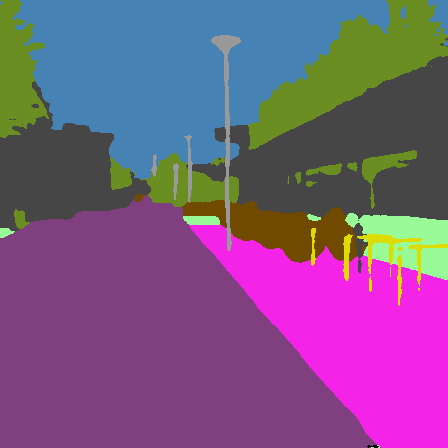}} \\
\\
& \multicolumn{1}{c}{(a)} & \multicolumn{1}{c}{(b)} & \multicolumn{1}{c}{(c)}  & \multicolumn{1}{c}{(d)}\\[6pt]
\rotatebox[origin=c]{90}{Input Image} & \raisebox{-0.5\height}{\includegraphics[width=\linewidth]{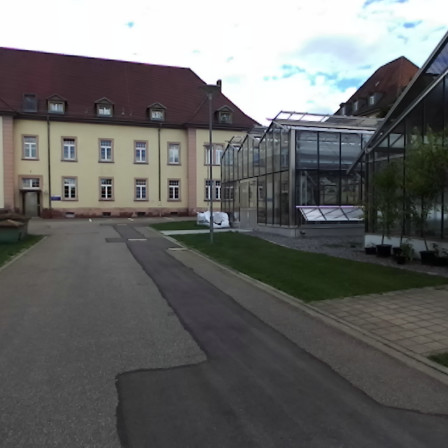}} & \raisebox{-0.5\height}{\includegraphics[width=\linewidth]{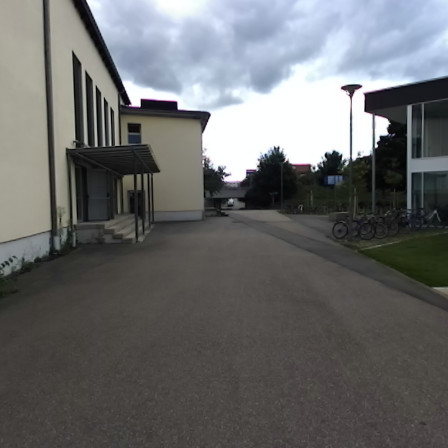}} & \raisebox{-0.5\height}{\includegraphics[width=\linewidth]{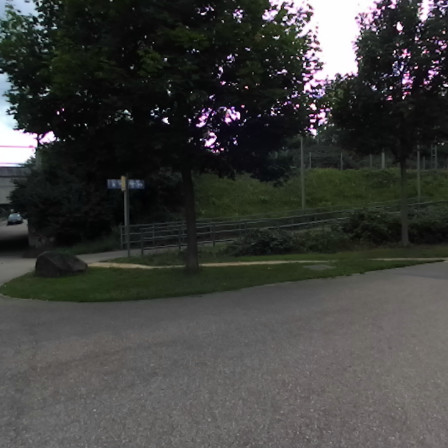}} & \raisebox{-0.5\height}{\includegraphics[width=\linewidth]{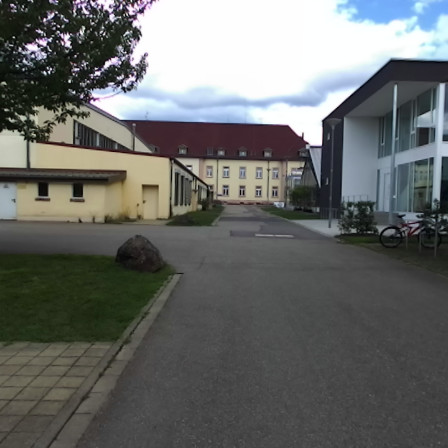}} \\
\\
\rotatebox[origin=c]{90}{Segmented Output} & \raisebox{-0.5\height}{\includegraphics[width=\linewidth]{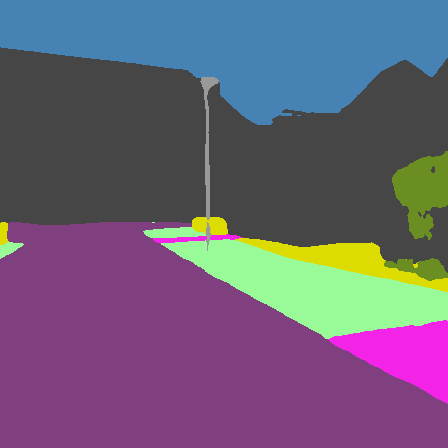}} & \raisebox{-0.5\height}{\includegraphics[width=\linewidth]{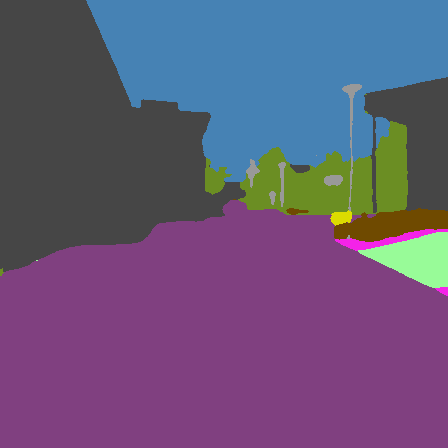}} & \raisebox{-0.5\height}{\includegraphics[width=\linewidth]{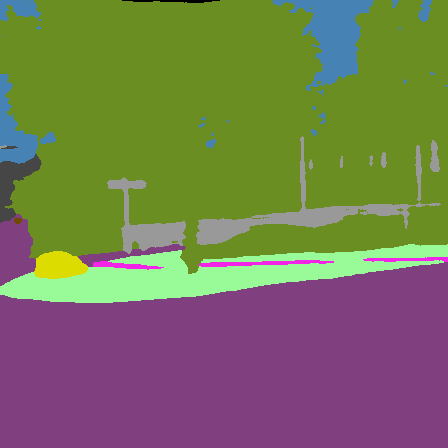}} & \raisebox{-0.5\height}{\includegraphics[width=\linewidth]{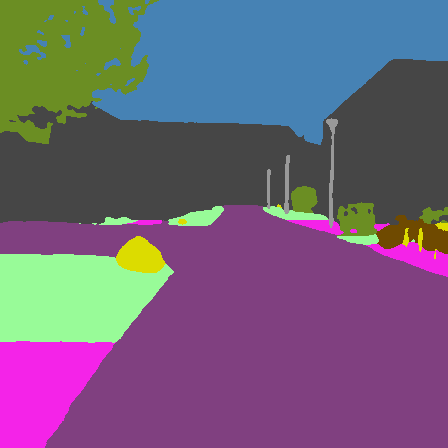}} \\
\\
& \multicolumn{1}{c}{(e)} & \multicolumn{1}{c}{(f)} & \multicolumn{1}{c}{(g)}  & \multicolumn{1}{c}{(h)}\\
\end{tabular} 
\caption{Qualitative evaluation of the segmentation results on the DeepLoc dataset in varying degrees of difficulty. The semantic categories are color coded as follows: \crule[sky]{0.2cm}{0.2cm}~Sky, \crule[road]{0.2cm}{0.2cm}~Road, \crule[sidewalk]{0.2cm}{0.2cm}~Sidewalk, \crule[grass]{0.2cm}{0.2cm}~Grass, \crule[vegetation]{0.2cm}{0.2cm}~Vegetation, \crule[building]{0.2cm}{0.2cm}~Building, \crule[pole]{0.2cm}{0.2cm}~Poles, \crule[dynamic]{0.2cm}{0.2cm}~Dynamic and \crule[other]{0.2cm}{0.2cm}~Other. VLocNet++ is able to accurately segment the scene while being robust to varying lighting conditions~(5(a, b, e)), shadows~(5(a)), reflective/translucent glass surfaces~(5(e, d, f, h)), thin and partially occluded structures (5(b, c, d, e)).}
\label{fig:qualitativeSeg}
\end{figure*}

In order qualitatively analyze the improvement in the segmentation due to the incorporation of the proposed adaptive warping technique, we performed experiments in comparison to the Adapnet~\citeNew{valada2017adapnet1} architecture that we build upon. \figref{fig:comparisonSeg} shows the qualitative results from this experiment on the DeepLoc dataset. \figref{fig:comparisonSeg}(a, c) show distant sidewalk paths in the middle of grass and structures partially occluded by vegetation, which are quite challenging to detect given the small input image resolution. While the segmentation output produced by Adapnet~\citeNew{valada2017adapnet1} often fails to capture the entire path/structure, VLocNet++ is able to accurately segment such categories. Similarly, \figref{fig:comparisonSeg}(b, d) include multiple thin structures such as poles and bike stands, which are either completely absent or only partially detected in the output of Adapnet~\citeNew{valada2017adapnet1}. By aggregating previous observations using the self-supervised warping scheme, our method is able to precisely segment thin pole-like structures. \figref{fig:comparisonSeg}(d) shows another challenging example, where a dark image produced by the camera due to direct sunlight, which causes grass to be misclassified as a bench by Adapnet~\citeNew{valada2017adapnet1}. While VLocNet++ is able to reliably distinguish between these categories even in the challenging condition.

Additionally, we show more qualitative segmentation results in challenging scenarios in~\figref{fig:qualitativeSeg}. \figref{fig:qualitativeSeg}(a, b, e, g) show varying lighting conditions that cause shadows, glare, over and under exposure due to sunlight. In all these cases, VLocNet++ yields an accurate representation of the scene overcoming these disturbances. Most of the scenes have thin pole-like structures such as lamp posts, sign posts, and fences that are the hardest to detect for any segmentation network. Although VLocNet++ utilizes a considerably small input image, it is still able to detect the entire structure of pole-like objects which can be seen in \figref{fig:qualitativeSeg}(b, c, d, g). This can be attributed to the dynamic warping scheme that we introduced that enables aggregation of information from previous observations. One of the challenging aspects of our dataset is the presence of glass buildings that are reflective or translucent. Despite the presence of several glass-constructs (\figref{fig:qualitativeSeg}(d, e, f, h)), our method is able to assign them to their correct semantic class. Another challenging aspect is to identify the boundary between grass and vegetation, which can be often challenging even for humans. \figref{fig:qualitativeSeg}(a, g) show examples where our network precisely predicts this boundary. Moreover, \figref{fig:qualitativeSeg}(e, g) show images with narrow paths surrounded by vegetation which are difficult to observe as the distance from the path increases. Inspecting the segmentation outputs for those cases, we can see that our model is able to capture such distant thin passages. 

\section{Additional Ablation Studies}
\label{sec:ablation}

In this section, we report additional ablation studies on the design choices for the various components of our VLocNet++ architecture.

\subsection{Base Architecture Topology}

In~\tabref{tab:ablation}, we present the different variants of the base architecture and activation function used along with the median localization error reported for each variant on the DeepLoc dataset. Models M1 to M3 employ shallow to deeper residual architectures with the standard ReLU activation function. The M3 model consisting of the ResNet-50 architecture~\citeNew{he2016deep} as the backbone shows an improved performance compared to the shallower ResNet-34 and ResNet-18 architectures. Moreover, the full preactivation ResNet-50 architecture~\citeNew{he2016identity} further improves upon the performance as it reduces overfitting and improves the convergence of the network. Furthermore, in the M5 model, we replace the ReLU activation with ELUs as they are more robust to noisy data and further accelerate the training. We denote this M5 model as our single-task VLocNet$++_{\text{STL}}$ architecture.

\begin{table}
\footnotesize 
\centering
\caption{Comparison of the VLocNet++ base architecture topology on the DeepLoc dataset.}
\label{tab:ablation}
\begin{tabular}{p{2.6cm}p{1.6cm}p{1cm}p{1.5cm}}
\noalign{\smallskip}\hline\noalign{\smallskip}
Method & Base Model & Activation & Median Error \\
\noalign{\smallskip}\hline\hline\noalign{\smallskip}
M1 & ResNet-18 & ReLU & $0.83\meter,\, 5.96\degree$ \\
M2 & ResNet-34 & ReLU & $0.57\meter,\, 4.04\degree$ \\
M3 & ResNet-50 & ReLU & $0.65\meter,\, 2.87\degree$ \\
M4 & PA ResNet-50 & ReLU & $0.57\meter,\, 2.44\degree$ \\
M5 (VLocNet$++_{\text{STL}}$) & PA ResNet-50 & ELU & $\mathbf{0.37\meter,\, 1.93\degree}$ \\
\noalign{\smallskip}\hline\noalign{\smallskip}
\end{tabular}
\end{table}

\subsection{Where to Warp?}

We conducted experiments to determine the network stage at which warping of feature maps from previous timesteps is effective. \tabref{tab:warpAblation} shows the mIoU achieved by warping the feature maps at different stages of the semantic segmentation stream. We hypothesize that warping feature maps at the end of a residual block before the next downsampling stage would be the most effective as the representations at the end of the block are more refined than at the beginning right after the downsampling. This is corroborated by the results shown in~\tabref{tab:warpAblation}, where warping feature maps at \textit{Res-3d} yields an increased mIoU than at \textit{Res-3a}. We also experimented with warping at multiple downsampling stages, however this only marginally increases the mIoU score. The highest improvement is obtained by warping at both \textit{Res-4f} and \textit{Res-5c} yielding an increase of $1.85\%$ in the mIoU score. Moreover, we can observe that warping at later residual blocks yields more improvement than at the earlier blocks.

\begin{table}
\footnotesize 
\centering
\caption{Improvement in the segmentation performance due to warping feature maps from the previous timestep. The warping layer denotes where the warping is performed in the segmentation stream. The results are shown on the DeepLoc dataset.}
\label{tab:warpAblation}
\begin{tabular}{p{3.5cm}p{1.6cm}}
\noalign{\smallskip}\hline\noalign{\smallskip}
Warping Layer & mIoU \\
\noalign{\smallskip}\hline\hline\noalign{\smallskip}
No warping & $78.59\%$ \\
\noalign{\smallskip}\hline\noalign{\smallskip}
\textit{Res-3a} & $80.03\%$ \\
\textit{Res-3d} & $80.19\%$ \\
\noalign{\smallskip}\hline\noalign{\smallskip}
\textit{Res-2c, Res-3d} & $80.09\%$ \\
\textit{Res-3d, Res-5c} & $80.34\%$ \\
\textit{Res-4f, Res-5c} & $\mathbf{80.44}\%$ \\
\noalign{\smallskip}\hline\noalign{\smallskip}
\textit{Res-3d, Res-4f, Res-5c} & $80.31\%$ \\
\noalign{\smallskip}\hline\noalign{\smallskip}
\end{tabular}
\end{table}

\subsection{Where to Fuse Semantic Features?}

In order to identify the stage where fusing semantic feature maps into the localization stream is most beneficial, we performed experiments fusing the semantic features at various units of the \textit{Res-4} block. Both \textit{Res-4} and \textit{Res-5} blocks have the same dimension as the semantic feature maps, however the units of the \textit{Res-5} block have a substantially large number of feature channels that would outweigh the rich semantic features. Therefore, we experiment with fusing with the units of the \textit{Res-4} block of the localization network. Results from this experiment shown in \tabref{tab:locFusion} demonstrates that the best performance is obtained by fusing the semantic feature maps at \textit{Res-4c} of the localization stream. Note that we do not experiment with fusing at \textit{Res-4b} as we fuse location-specific feature maps from this layer into the segmentation stream and adding this connection would result in a cyclic dependency between the respective network streams.

\begin{table}
\footnotesize 
\centering
\caption{Improvement in the localization performance due to the fusion of semantic feature maps. The fusion layer denotes where the semantic feature maps are fused into the localization stream. The results are shown on the DeepLoc dataset.}
\label{tab:locFusion}
\begin{tabular}{p{2cm}p{2.2cm}p{2cm}}
\noalign{\smallskip}\hline\noalign{\smallskip}
Fusion Layer & Median & Median \\
 & Translational Error & Rotational Error \\
\noalign{\smallskip}\hline\hline\noalign{\smallskip}
No fusion & $0.37\meter$ & $1.93\degree$ \\
\noalign{\smallskip}\hline\noalign{\smallskip}
\textit{Res-4a} & $0.49\meter$ & $3.10\degree$ \\
\textit{Res-4c} & $\mathbf{0.32\meter}$ & $\mathbf{1.48\degree}$ \\
\textit{Res-4d} & $0.54\meter$ & $1.30\degree$ \\
\textit{Res-4e} & $0.46\meter$ & $1.95\degree$ \\
\textit{Res-4f} & $0.61\meter$ & $1.45\degree$ \\
\noalign{\smallskip}\hline\noalign{\smallskip}
\end{tabular}
\end{table}

\subsection{Baseline Multitask Architectures}

\begin{table}
\footnotesize 
\centering
\caption{Comparison of VLocNet$++_{\text{MTL}}$ with baseline models for fusing semantic features into the localization stream. Results are shown for the DeepLoc dataset.}
\label{tab:ablationMTL}
\begin{tabular}{p{3cm}p{2.2cm}p{2cm}}
\noalign{\smallskip}\hline\noalign{\smallskip}
Method & Median & Median \\
 & Translational Error & Rotational Error \\
\noalign{\smallskip}\hline\hline\noalign{\smallskip}
VLocNet$++_{\text{STL}}$ & $0.37\meter$ & $1.93\degree$ \\
\noalign{\smallskip}\hline\noalign{\smallskip}
MTL\_input-concat & $0.56\meter$ & $3.63\degree$ \\
MTL-mid-concat \textit{Res-4c} & $0.55\meter$ & $3.38\degree$ \\
MTL-mid-concat \textit{Res-4f} & $0.50\meter$ & $3.10\degree$ \\
MTL-shared & $1.17\meter$ & $4.20\degree$ \\
\noalign{\smallskip}\hline\noalign{\smallskip}
VLocNet$++_{\text{MTL}}$ & $\mathbf{0.32\meter}$ & $\mathbf{1.48\degree}$ \\
\noalign{\smallskip}\hline\noalign{\smallskip}
\end{tabular}
\end{table}

In this section, we elaborate the topologies of the various baseline multitask architectures for encoding semantic features into the localization network stream. We compare the performance of VLocNet$++_{\text{MTL}}$ with
\begin{itemize}
\item \textbf{MTL\_input-concat}: A simple approach to incorporate the semantics into the localization stream is by concatenating the semantic segmentation output $M_t$ as a fourth channel to the input image $I_t$ and feeding the resulting four channel tensor as input to the localization stream. As shown in \tabref{tab:ablationMTL}, this drastically reduces the performance of the model in comparison to the single-task VLocNet$++_{\text{STL}}$.
\item \textbf{MTL\_mid-concat}: For the second baseline, we concatenate the semantic feature maps with intermediate feature maps of the localization stream. Our proposed VLocNet$++_{\text{MTL}}$ architecture fuses semantic feature maps at \textit{Res-4c} of the localization stream using our proposed fusion layer. In order to compare the effectiveness of our proposed fusion layer, we fuse semantic feature maps at \textit{Res-4c} of the localization stream using simple concatenation. As shown in \tabref{tab:ablationMTL}, this model achieves a slightly better performance than the MTL\_input-concat model but it is substantially worse than the single-task model. We also compare with fusing at \textit{Res-4f}, which still achieves a lower performance than VLocNet$++_{\text{STL}}$.
\item \textbf{MTL\_shared}: Finally, we compare with an approach~ \cite{abdulnabi2015multi} that shares the latent spaces of both the semantic segmentation and localization streams. This approach achieves the lowest performance as both these tasks significantly differ in the representations that they learn and sharing weights across both these network streams only lowers their individual task performances.
\end{itemize}

However, as shown in \tabref{tab:ablationMTL}, VLocNet$++_{\text{MTL}}$ substantially outperforms all the baselines by a large margin. More importantly, it outperforms the single-task VLocNet$++_{\text{STL}}$ model demonstrating the utility of fusing semantic features into the localization network using our proposed fusion layer.\balance

\footnotesize
\bibliographystyleNew{IEEEtran}
\bibliographyNew{references2}

\end{document}